\documentclass[10pt,twocolumn,letterpaper]{article}

\usepackage{amsmath,amsfonts,bm}

\def\eqref#1{equation~\ref{#1}}
\def\1{\bm{1}}

\def\rd{{\textnormal{d}}}

\def\rx{{\textnormal{x}}}
\def\ry{{\textnormal{y}}}

\def\rvx{{\mathbf{x}}}

\def\vw{{\bm{w}}}
\def\vx{{\bm{x}}}

\DeclareMathAlphabet{\mathsfit}{\encodingdefault}{\sfdefault}{m}{sl}
\SetMathAlphabet{\mathsfit}{bold}{\encodingdefault}{\sfdefault}{bx}{n}

\newcommand{\R}{\mathbb{R}}

\DeclareMathOperator*{\argmax}{arg\,max}

\usepackage{cvpr}
\usepackage{times}
\usepackage{epsfig}
\usepackage{graphicx}
\usepackage{amsmath}
\usepackage{amssymb}
\usepackage{booktabs} 
\usepackage{multirow}
\usepackage{subfig}
\usepackage{rotating}

\usepackage[breaklinks=true,bookmarks=false]{hyperref}

\iftrue
    \newcommand{\zk}[1]{}
    \newcommand{\yh}[1]{}
    \newcommand{\ys}[1]{}
\else
    \newcommand{\zk}[1]{{\color{blue}{(Zsolt: #1)}}}
    \newcommand{\yh}[1]{{\color{red}{(Yen-Chang: #1)}}}
    \newcommand{\ys}[1]{{\color{cyan}{(Yilin: #1)}}}
\fi

\cvprfinalcopy % *** Uncomment this line for the final submission

\def\cvprPaperID{9419} % *** Enter the CVPR Paper ID here

\ifcvprfinal\fi
\begin{document}

%%%%%%%%% TITLE
\title{Generalized ODIN: Detecting Out-of-distribution Image without Learning from Out-of-distribution Data}
%\title{Toward learning the conditional classifier for detecting out-of-distribution image}

\author{Yen-Chang Hsu\textsuperscript{1}, Yilin Shen\textsuperscript{2}, Hongxia Jin\textsuperscript{2}, Zsolt Kira\textsuperscript{1} \\
\normalsize
\textsuperscript{1}Georgia Institute of Technology,
\textsuperscript{2}Samsung Research America
}

\iffalse
\author{Yen-Chang Hsu\\
Gatech\\
Atlanta, USA\\
{\tt\small yenchang.hsu@gatech.edu}
\and
Yilin Shen, Hongxia Jin\\
Samsung Research America\\
Mountain View, USA\\
{\tt\small \{yilin.shen,hongxia.jin\}@samsung.com}
\and
Zsolt Kira\\
Gatech\\
Atlanta, USA\\
{\tt\small zkira@gatech.edu}
}
\fi

\maketitle
\thispagestyle{empty}

%%%%%%%%% ABSTRACT
\begin{abstract}
Deep neural networks have attained remarkable performance when applied to data that comes from the same distribution as that of the training set, but can significantly degrade otherwise. Therefore, detecting whether an example is out-of-distribution (OoD) is crucial to enable a system that can reject such samples or alert users. Recent works have made significant progress on OoD benchmarks consisting of small image datasets. However, many recent methods based on neural networks rely on training or tuning with both in-distribution and out-of-distribution data. The latter is generally hard to define \textit{a-priori}, and its selection can easily bias the learning. We base our work on a popular method ODIN\footnote{ODIN: Out-of-DIstribution detector for Neural networks \cite{liang2017enhancing}} \cite{liang2017enhancing}, proposing two strategies for freeing it from the needs of tuning with OoD data, while improving its OoD detection performance. We specifically propose to decompose confidence scoring as well as a modified input pre-processing method. We show that both of these significantly help in detection performance. Our further analysis on a larger scale image dataset shows that the two types of distribution shifts, specifically semantic shift and non-semantic shift, present a significant difference in the difficulty of the problem, providing an analysis of when ODIN-like strategies do or do not work.
\end{abstract}

%%%%%%%%% BODY TEXT
\section{Introduction}

State-of-the-art machine learning models, specifically deep neural networks, are generally designed for a static and closed world. The models are trained under the assumption that the input distribution at test time will be the same as the training distribution. In the real world, however, data distributions shift over time in a complex, dynamic manner. Even worse, new concepts (\eg new categories of objects) can be presented to the model at any time. Such within-class distribution shift and unseen concepts both may lead to catastrophic failures since the model still attempts to make predictions based on its closed-world assumption. These failures are therefore often silent in that they do not result in explicit errors in the model.

The above issue had been formulated as a problem of detecting whether an input data is from in-distribution (\ie the training distribution) or out-of-distribution (\ie a distribution different from the training distribution) \cite{hendrycks2016baseline}. This problem has been studied for many years \cite{hellman1970nearest} and has been discussed in several views such as rejection \cite{geifman2017selective,CortesRejection}, anomaly detection \cite{andrews2016transfer}, open set recognition \cite{bendale2015towards}, and uncertainty estimation \cite{MalininNIPS2018, malinin2019reverse, malinin2019ensemble}.\yh{updated} In recent years, a popular neural network-based baseline is to use the max value of class posterior probabilities output from a softmax classifier, which can in some cases be a good indicator for distinguishing in-distribution and out-of-distribution inputs \cite{hendrycks2016baseline}. 

ODIN \cite{liang2017enhancing}, based on a trained neural network classifier, provides two strategies, temperature scaling and input preprocessing, to make the max class probability a more effective score for detecting OoD data. Its performance has been further confirmed by \cite{Shafaei2019}, where 15 OoD detection methods are compared with a less biased evaluation protocol. ODIN out-performs popular strategies such as MC-Dropout \cite{gal2016dropout}, DeepEnsemble \cite{lakshminarayanan2017simple}, PixelCNN++ \cite{salimans2017pixelcnn}, and OpenMax \cite{bendale2016towards}.  

Despite its effectiveness, ODIN has a requirement that it needs OoD data to tune hyperparameters for both its strategies, leading to a concern that hyperparameters tuned with one out-of-distribution dataset might not generalize to others, discussed in \cite{Shafaei2019}. In fact, other neural network-based methods \cite{lee2018simple, vyas2018out}, which follow the same problem setting, have a similar requirement. \cite{Dhamija18, hendrycks2018deep} push the idea of utilizing OoD data further by using a carefully chosen OoD dataset to regularize the learning of class posteriors so that OoD data have much lower confidence than in-distribution. Lastly, \cite{lee2017training} uses a generative model to generate out-of-distribution data around the boundary of the in-distribution for learning.

Although the above works show that learning with OoD data is effective, the space of OoD data (ex: image pixel space) is usually too large to be covered, potentially causing a selection bias for the learning. Some previous works have done a similar attempt to learn without OoD data, such as \cite{shalev2018out}, which uses word embeddings for extra supervision, and \cite{masana2018metric} which applies metric learning criteria. However, both works report performance similar to ODIN, showing that learning without OoD data is a challenging setting.

\begin{figure}
  \centering
  \includegraphics[clip, trim=0cm 8.2cm 12.2cm 0cm, width=0.47\textwidth]{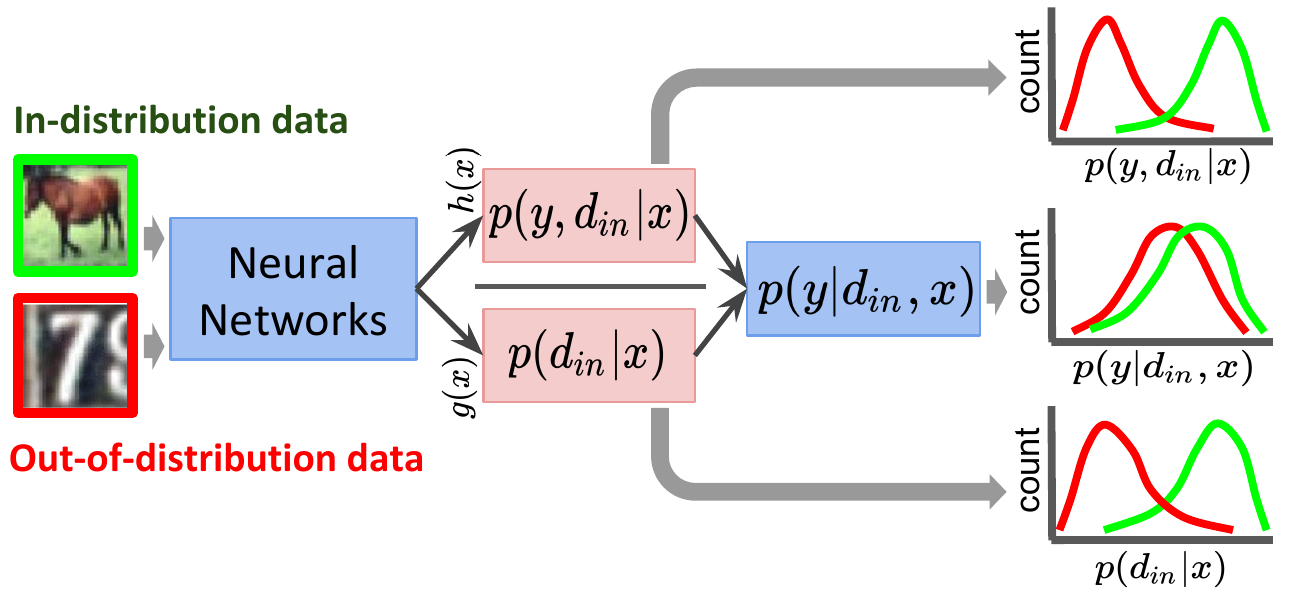} 
  \caption{The concept of detecting out-of-distribution images by encouraging neural networks to output scores, $h(x)$ and $g(x)$, to behave like the decomposed factors in the conditional probability when the close-world assumption $d_{in}$ is explicitly considered. Its elucidation is in Section \ref{sec:deconf}. A small overlap between the green and red histograms means the x-axis a good scoring function for distinguishing OoD data from in-distribution. The extent of overlap is usually measured by AUROC, elaborated in Section \ref{sec:exp_setting}.\yh{updated} }\label{fig:deconf_overview}
\end{figure}

In this work, we closely follow the setting of ODIN, proposing two corresponding strategies for the problem of learning without OoD data. First, we provide a new probabilistic perspective for decomposing confidence of predicted class probabilities. We specifically add a variable for explicitly adopting the closed world assumption, representing whether the data is in-distribution or not, and discuss its role in a decomposed conditional probability. Inspired by the probabilistic view, we use a dividend/divisor structure for a classifier, which encourages neural networks to behave similarly to the decomposed confidence effect. The concept is illustrated in Figure \ref{fig:deconf_overview}, and we note the dividend/divisor structure is closely related to temperature scaling except that the scale depends on the input instead of a tuned hyperparameter. Second, we build on the input preprocessing method from ODIN \cite{liang2017enhancing} and develop an effective strategy to tune its perturbation magnitude (which is a hyperparameter of the preprocessing method) with only in-distribution data. 

We then perform extensive evaluations on benchmark image datasets such as CIFAR10/100, TinyImageNet, LSUN, SVHN, as well as a larger scale dataset DomainNet, for investigating the conditions under which the proposed strategies do or do not work. The results show that the two strategies can significantly improve upon ODIN, achieving a performance close to, and in some cases surpassing, state-of-the-art methods \cite{lee2018simple} which use out-of-distribution data for tuning. Lastly, our systematical evaluation with DomainNet reveals the relative difficulties between two types of distribution shift: semantic shift and non-semantic shift, which are defined by whether a shift is related to the inclusion of new semantic categories.

In summary, the contribution of this paper is three-fold:
\begin{itemize}
	\item A new perspective of decomposed confidence for motivating a set of classifier designs that consider the closed-world assumption.\yh{updated}
	\item A modified input preprocessing method without tuning on OoD data.
	\item Comprehensive analysis with experiments under the setting of learning without OoD data. 
\end{itemize}
%------------------------------------------------------------------------
\section{Background} \label{sec:background}

\begin{figure}
  \centering
  \includegraphics[clip, trim=0cm 5cm 6cm 0cm, width=0.5\textwidth]{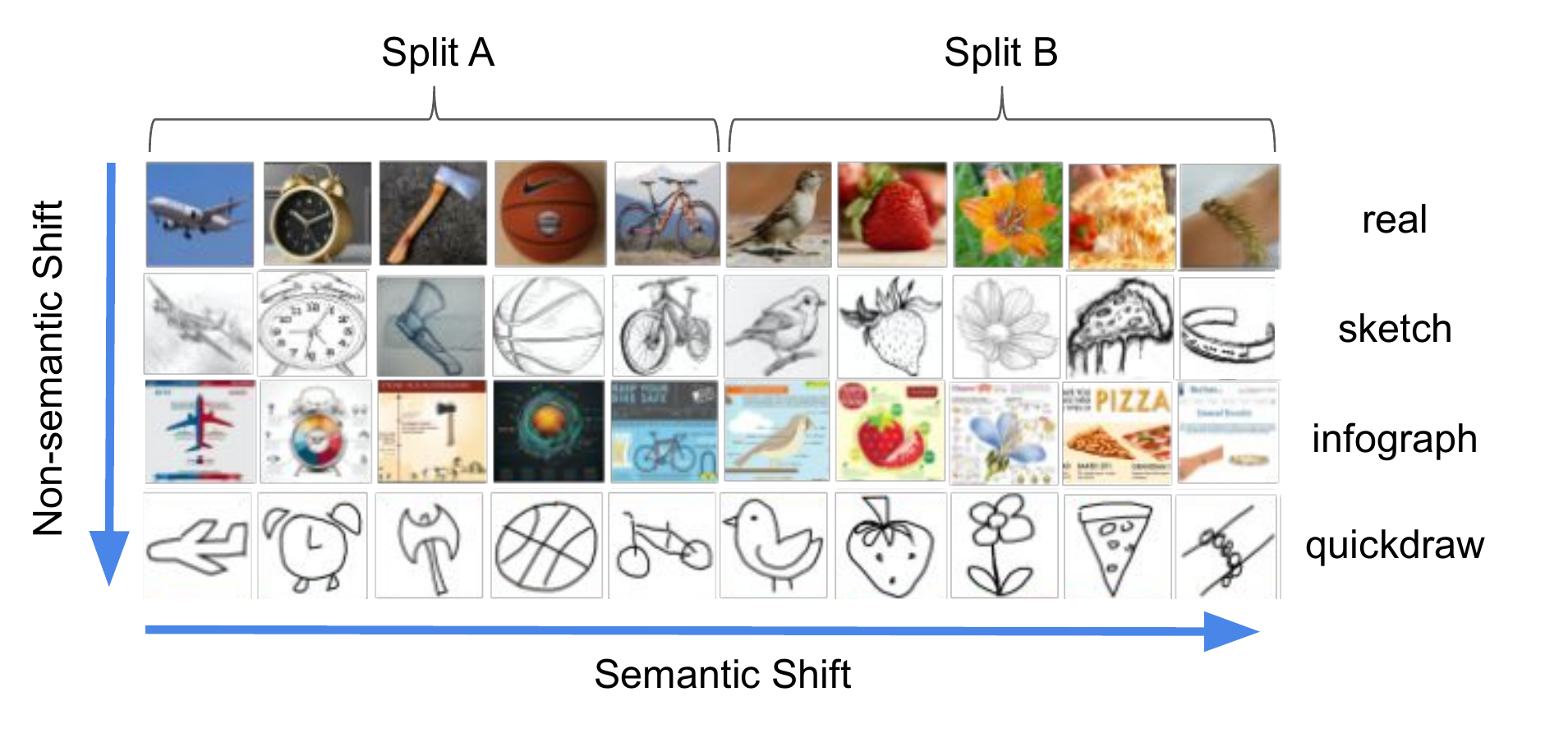} 
  \caption{An example scheme of semantic shift and non-semantic shift. It is illustrated with DomainNet \cite{peng2018moment} images. The setting with two splits (A and B) will be used in our experiments, where only real-A is the in-distribution data.}\label{fig:shifts}
\end{figure}

This work considers the OoD detection setting in classification problems. We begin with a dataset $D_{in}=\{(\vx_i, y_i)\}^N_{i=1}$, denoting in-distribution data $\vx_i \in \R^k$ and categorical label $y_i \in \{\ry\} = \{1..C\}$ for $C$ classes. $D_{in}$ is generated by sampling from a distribution $p_{in}(\rvx, \ry)$. We then have a discriminative model $f_{\theta}(\vx)$ with parameters $\theta$ learned with the in-domain dataset $D_{in}$, predicting the class posterior probability $p(\ry|\vx)$.

When the learned classifier $f_{\theta}$ is deployed in the open world, it may encounter data drawn from a different distribution $p_{out}$ such that $p_{out} \neq p_{in}$. Sampling from all possible distributions $p_{out}$ that might be encountered is generally intractable especially when the dimension $k$ is large, such as in the cases of image data. Note also that we can conceptually categorize the type of differences into non-semantic shift and semantic shift. Data with non-semantic shift is drawn from the distribution $p_{out}(\rvx, \ry)$. Examples with this shift come from the same object class but are presented in different forms, such as cartoon or sketch images. Such shift is also a scenario be widely discussed in the problem of domain adaptation \cite{patel2015visual, peng2018moment}. In the case of semantic shift, the data is drawn from a distribution $p_{out}(\rx, \bar\ry)$ with $\{\bar\ry\} \cap \{\ry\}=\emptyset$. In other words, the data is from a class not seen in the training set $D_{in}$. Figure \ref{fig:shifts} has an illustration. 

The above separation leads to two natural questions that must be answered for a model to work in an open world: How can the model avoid making a prediction when encountering an input $\vx \sim p_{out}(\rx, \bar\ry)$, or reject a low confidence prediction when $\vx \sim p_{out}(\rx, \ry)$? 
In this work, we propose to introduce an  \textit{explicit} binary domain variable $\rd \in \{d_{in},d_{out}\}$ in order to represent this decision, with $d_{in}$ meaning that the input is $\vx \sim p_{in}$ while $d_{out}$ meaning $\vx \nsim p_{in}$ (or equivalently $\vx \sim p_{out}$). Note that while generally the model cannot distinguish between the two cases we defined, we can still show that both of the questions above can be answered by estimating this single variable $\rd$.

The ultimate goal, then, is to find a scoring function $S(\vx)$ which  correlates to the domain posterior probability $p(\rd|\vx)$, in that a higher score $s$ from $S(\vx)$ indicates a higher probability of $p(d_{in}|\vx)$. The binary decision now can be made by applying a threshold on $s$. Selecting such a  threshold is subject to the application requirement or the performance metric calculation protocol. With the above notation, we can view the baseline method \cite{hendrycks2016baseline} as a special case with a specific scoring function $S_{Base}(\vx) = \max_{\ry} p(\ry|\vx)$, where $p(\ry|\vx)$ is obtained from a standard neural network classifier $f_\theta$ trained with cross-entropy loss. However, $S(\vx)$ can become a learnable parameterized function, and different OoD methods can then be categorized by specific parameterizations and learning procedures. A key differentiator between methods is whether the parameters are learned with or without OoD data.

\subsection{Related Methods} \label{sec:odin_maha} 

This section describes the two methods that are the most related to our work: ODIN \cite{liang2017enhancing} and Mahalanobis \cite{lee2018simple}. These two methods will serve as strong baselines in our evaluation, especially since Mahalanobis has further been shown to have significant advantages over ODIN. Note that both ODIN and Mahalanobis start from a vanilla classifier $f_{\theta}$ trained on $D_{in}$, then have a scoring function $S(\vx; f_{\theta})$ which has extra parameters to be tuned. In their original work, those parameters are specifically tuned for each OoD dataset. Here we will describe methods to use them without tuning on OoD data.

\textbf{ODIN} comprises two strategies: temperature scaling and input preprocessing. The temperature scaling is applied to its scoring function, which has $f_i(\vx)$ for the logit of $i$ class:
\begin{equation} \label{eq:ODIN}
S_{ODIN}(\vx)=\max_i \frac{\exp{(f_i(\vx)/T)}}{\sum_{j=1}^C{\exp{(f_j(\vx)/T)}}}
\end{equation}

Although ODIN originally involved tuning the hyperparameter $T$ with out-of-distribution data, it was also shown that a large $T$ value can generally be preferred, suggesting that the gain is saturated after 1000 \cite{liang2017enhancing}. We follow this guidance and fix $T=1000$ in our experiments.

\textbf{Mahalanobis} comprises two parts as well: Mahalanobis distance calculation and input preprocessing. The score is calculated with Mahalanobis distance as follows:
\begin{gather} \label{eq:Maha}
S^{\ell}_{Maha}(\vx)=\max_i -(f^{\ell}(\vx) - \mu^{\ell}_i)^T \Sigma_{\ell}^{-1} (f^{\ell}(\vx) - \mu^{\ell}_i), \hspace{-.5em}\\
S_{Maha}(\vx)=\sum_{\ell}\alpha_{\ell} S^{\ell}_{Maha}(\vx)
\end{gather}

The $f^{\ell}(\vx)$ represents the output features at the $\ell$th-layer of neural networks, while $\mu_i$ and $\Sigma$ are the class mean representation and the covariance matrix, correspondingly. The hyperparameter is $\alpha_{\ell}$. In the original method, $\alpha_{\ell}$ is regressed with a small validation set containing both in-distribution and out-of-distribution data. Therefore they have a set of $\alpha_{\ell}$ tuned for each OoD dataset. As a result, for the baseline that does not tune on OoD data we use uniform weighting $S_{Maha}(\vx)=\sum_{\ell} S^{\ell}_{Maha}(\vx)$. 

Note that both methods use the input preprocessing strategy, which has a hyperparameter to be tuned. In their original works, this hyperparameter is tuned for each OoD dataset as well. Therefore we develop a version that does not require tuning with out-of-distribution data.% to provide a strong baseline that matches our setting.

%-------------------------------------------------------------------------
\section{Approach} \label{sec:approach}
\subsection{The Decomposed Confidence} \label{sec:deconf}
\cite{szegedy2013intriguing, nguyen2015deep, hendrycks2016baseline} observed that the softmax classifier tends to output a highly confident prediction, reporting that "random Gaussian noise fed into an MNIST image classifier gives a predicted class probability of 91\%". They attribute this to the use of the softmax function which is a smooth approximation of an indicator function, hence tending to give a spiky distribution instead of a uniform distribution over classes \cite{hendrycks2016baseline}. We acknowledge this view and further consider it as a limitation in the design of the softmax classifier. To address this limitation, our inspiration starts from reconsidering its outputs, the class posterior probability $p(\ry|\vx)$, which does not consider the domain $\rd$ at all. In other words, current methods condition on domain $\rd=d_{in}$ based on the implicit closed world assumption. Thus, we use our explicit variable $d_{in}$ in the classifier, rewriting it as the quotient of the joint class-domain probability and the domain probability using the rule of conditional probability:
\begin{equation} \label{eq:deconf}
%p(\ry|d=1, \vx) = \frac{p(\ry, \rd=1,\vx)}{p(\rd=1,\vx)}
%=  \frac{p(\ry, \rd=1|\vx)p(\vx)}{p(\rd=1|\vx)p(\vx)}
%=  \frac{p(\ry, \rd=1|\vx)}{p(\rd=1|\vx)}
p(\ry|d_{in}, \vx) = \frac{p(\ry, d_{in}|\vx)}{p(d_{in}|\vx)}
\end{equation}

Equation \ref{eq:deconf} provides a probabilistic view of why classifiers tend to be overconfident. Consider an example $\vx \sim p_{out}$: It is natural to expect that the joint probability $P(\ry, d_{in}|\vx)$ is low (e.g. $0.09$) for its maximum value among C classes. One would also expect its domain probability $p(d_{in}|\vx)$ is low (e.g. $0.1$). Therefore, calculating $p(\ry|d_{in}, \vx)$ with Equation \ref{eq:deconf} gives a high probability ($0.9$), demonstrating how overconfidence can result. Based on the form of Equation \ref{eq:deconf}, we call  $p(\ry, d_{in}|\vx)$ and $p(d_{in}|\vx)$ the decomposed confidence scores.

One straightforward solution for the above issue is to learn a classifier to predict the joint probability $p(\ry, d_{in}|\vx)$ by having both supervision on class $\ry$ and domain $\rd$. Learning to predict $p(\ry, d_{in}|\vx)$ is preferred over $p(d_{in}|\vx)$ because it can serve both purposes for predicting a class by $\argmax_{y_{in}}p(\ry, d_{in}|\vx)$ and rejecting a prediction by thresholding. This idea relates to the work of \cite{hendrycks2018deep}, which adds an extra loss term to penalize a predicted non-uniform class probability when an out-of-distribution data is given to the classifier. However, this strategy requires out-of-distribution data for regularizing the training.

Without having supervision on domain $\rd$ (\ie without out-of-distribution data), there is no principled way to learn $p(\ry, d_{in}|\vx)$ and $p(d_{in}|\vx)$. This situation is similar to unsupervised learning (or self-supervised learning) in that we need to insert assumptions or prior knowledge about the task for learning. In our case, we use the dividend/divisor structure in Equation \ref{eq:deconf} as the prior knowledge to design the structure of classifiers, providing classifiers a capacity to decompose the confidence of class probability. 

In the dividend/divisor structure for classifiers, we define the logit $f_i(\vx)$ for class $i$, which is the division between two functions $h_i(\vx)$ and $g(\vx)$:
\begin{equation} \label{eq:deconf2}
f_i(\vx)=\frac{h_i(\vx)}{g(\vx)},%, \text{ with }  p(\ry=i|d_{in}, \vx) = \frac{\exp{f_i(\vx)}}{\sum_{j=1}^C{\exp{f_j(\vx)}}}. 
\end{equation}
The quotient $f_i(x)$ is then normalized by the exponential function (\ie softmax) for outputting a class probability $p(\ry=i|d_{in}, \vx)$, which is subject to cross-entropy loss. 

With the exponential normalization effect of softmax, the cross-entropy loss can be minimized in two ways: increasing $h_i(\vx)$ or decreasing $g(\vx)$. In other words, when the data is not in the high-density region of in-distribution, $h_i(\vx)$ may tend towards smaller values. In such case, the $g(\vx)$ is encouraged to be small so that the resulting logits $f_i(\vx)$ can further minimize the cross-entropy loss. In the other case when the data is in the high density region, $h_i(\vx)$ generally can reach a higher value relatively easier, thus its corresponding $g(\vx)$ value is less encouraged to go small. The discussed interaction between $h_i(\vx)$ and $g(\vx)$ is the primary driving force to encourage $h_i(\vx)$ to behave similarly to $p(\ry=i, d_{in}|\vx)$ and $g(\vx)$ to behave similarly to $p(d_{in}|\vx)$, in a way that the distributional overlap between the scores of OoD and in-distribution data is small, which is an intrinsic property of $p(y,d_{in}|\vx)$ and $p(d_{in}|\vx)$, illustrated in Figure 1.\yh{update}

\subsubsection{Design Choices}

Although the dividend/divisor structure provides a tendency, it does not necessarily guarantee the decomposed confidence effect to happen. The characteristic of $h_i(\vx)$ and $g(\vx)$ can largely affect how likely the decomposition could happen. Therefore we discuss a set of simple design choices to investigate whether such decomposition is generally obtainable. 

Specifically we have $g(\vx)=\sigma(BN(\vw_gf^p(\vx) + b_g))$, which uses features $f^p(\vx)$ from the penultimate layer of neural networks sequentially through another linear layer, batch normalization ($BN$, optional for a faster convergence), and a sigmoid function $\sigma$. The $\vw$ and $b$ represent the learnable weights. For $h_i(\vx)$, we investigate three similarity measures, including inner-product (I), negative Euclidean distance (E), and cosine similarity (C) for $h_i^I(\vx)$, $h_i^E(\vx)$, and $h_i^C(\vx)$, correspondingly:
\begin{align} \label{eq:hi}
h_i^I(\vx)&=\vw_i^T f^p(\vx) + b_i;\\
h_i^E(\vx)&=-\lVert f^p(\vx)-\vw_i \rVert^2; \\
h_i^C(\vx)&=\frac{\vw_i^T f^p(\vx)}{\lVert \vw_i \rVert  \lVert f^p(\vx) \rVert}
\end{align}

The overall neural network model $f_{\theta}$ therefore has two branches ($h_i$ and $g$) after its penultimate layer (See Figure \ref{fig:deconf_overview}). At training time, the model calculates the logit $f_i$ followed by the softmax function with cross-entropy loss on top of it. At testing time, the class prediction can be made by either calculating $\argmax_i f_i(\vx)$ or $\argmax_i h_i(\vx)$ (both will give the same predictions). For out-of-distribution detection, we use the scoring function $S_{DeConf}(\vx)=\max_i{h_i(\vx)}$ or $g(\vx)$. 

Note that when $h_i(\vx)=h_i^I(\vx)$ and $g(\vx)=1$, this method reduces to the baseline \cite{hendrycks2016baseline}. We call the three variants of our method DeConf-I, DeConf-E, and DeConf-C. For simplicity, the above names represent using $h_i(\vx)$ for the scores. The use of $g(\vx)$ will be indicated specifically.

\subsubsection{Temperature Scaling}

The $g(x)$ in Equation \ref{eq:deconf2} can be immediately viewed as a learned temperature scaling function discussed in \cite{neumann2018relaxed} and a concurrent report \cite{OOD_temp_cosine}.\yh{updated} However, our experiment results strongly suggest that $g(x)$ is more than a scale. The $g(x)$ achieves an OoD detection performance significantly better than baselines in many experiments, indicating its potential in estimating the $p(d_{in}|\vx)$.\yh{updated} More importantly, the temperature scaling is generally used as a numerical trick for learning a better embedding \cite{zhang2019adacos}, softening the prediction \cite{hinton2015distilling}, or calibrating the confidence \cite{guo2017calibration}. Our work provides a probabilistic view for its effect, indicating such temperature might relate to how strong a classifier assumes a closed world as a prior.

\subsection{A Modified Input Preprocessing Strategy} \label{sec:IPP}

This section describes a modified version of the input preprocessing method proposed in ODIN \cite{liang2017enhancing}. The primary purpose of the modification is making the search of the perturbation magnitude $\epsilon$ to not rely on out-of-distribution data. The perturbation of input is given by:
\begin{equation}
\hat{\vx}=\vx-\epsilon \text{sign}(-\nabla_\vx S(\vx)) 
\end{equation}

In the original method \cite{liang2017enhancing} the best value of $\epsilon$ is searched with a half-half mixed validation dataset of $D_{in}^{val} \sim p_{in}$ and $D_{out}^{val} \sim p_{out}$ over a list of 21 values. The perturbed images $\hat{\vx}$ are fed into the classification model $f_\theta$ for calculating the score $S(\vx)$. The performance of each magnitude is evaluated with the benchmark metric  (TNR@TPR95, described later) and the best one is selected. This process repeats for each out-of-distribution dataset, and therefore the original method results in a number of $\epsilon$ values equal to the number of out-of-distribution datasets in the benchmark.

In our method, we search for the $\epsilon^*$ which maximizes the score $S(\vx)$ with only the in-distribution validation dataset $D_{in}^{val}$:
\begin{equation} \label{eq:epsilon}
\epsilon^* = \argmax_{\epsilon} \sum_{x \in D_{in}^{val}}S(\hat{\vx})
\end{equation}

Our searching criteria is still based on the same observation made by \cite{liang2017enhancing}. They observe that the in-distribution images tend to have their score $s$ increased more than the out-of-distribution images when the input perturbation is applied. We therefore use Eq. \ref{eq:epsilon} since we argue that an $\epsilon$ which makes a large score increase for in-distribution data should be sufficient to create a distinction in score. Our method also does not even require class labels although it is available in $D_{in}^{val}$. More importantly, our method selects only one $\epsilon$ based on $D_{in}^{val}$ without access to the benchmark performance metric (\eg TNR@TPR95), greatly avoiding the hyperparameter from fitting to a specific benchmark score. Lastly, we perform the search of $\epsilon$ on a much coarser grid, which has only 6 values: $[0.0025, 0.005, 0.01, 0.02, 0.04, 0.08]$. Therefore, our search is much faster. Although overshooting is possible (\eg the maximum value is at the middle of two scales in the grid) due to the coarser grid, it can be mitigated by reducing the found magnitude by one scale (\ie divide it by two). This simple strategy consistently gains or maintains the performance on varied scoring functions, such as $S_{Base}$, $S_{DeConf}$, $S_{ODIN}$, and $S_{Maha}$.

The method in this section is orthogonal to all the methods evaluated in this work. For convenience, we will add a * after the name of other methods to indicate a combination, for example, Baseline* and DeConf-C*.

\section{Experiments}

\subsection{Experimental Settings} \label{sec:exp_setting}

\textbf{Overall procedure:} In all experiments, we first train a classifier $f_{\theta}$ on an in-distribution training set, then tune the hyperparameters (\eg the perturbation magnitude $\epsilon$) on an in-distribution validation set without using its class labels. At testing time, the OoD detection scoring function $S(\vx)$ calculates the scores $s$ from the outputs of $f_{\theta}$. The scores $s$ is calculated for both in-distribution validation set $D_{in}^{val}$ and out-of-distribution dataset $D_{out} \sim p_{out}$. The scores $s$ are then sent to a performance metric calculation function. The above procedure is the same as related works in this line of research \cite{liang2017enhancing, lee2018simple, hendrycks2018deep, Shafaei2019, vyas2018out, lee2017training}, except that we do not use OoD data for tuning the hyperparameters in the scoring function $S(\vx)$.

\textbf{In-distribution Datasets:} We use SVHN \cite{netzer2011reading} and CIFAR-10/100 images with size 32x32 \cite{krizhevsky2009learning} for the classification task. Detecting OoD with CIFAR-100 classifier is generally harder than CIFAR-10 and SVHN, since a larger amount of classes usually involves a wider range of variance, and thus it has a higher tendency to treat random data (\eg Gaussian noise) as in-distribution. For that reason, we use CIFAR-100 in our ablation and robustness study.

\textbf{Out-of-distribution Datasets:} We include all the OoD datasets used in ODIN \cite{liang2017enhancing}, which are TinyImageNet(crop), TinyImageNet(resize), LSUN(crop), LSUN(resize), iSUN, Uniform random images, and Gaussian random images. We further add SVHN, a colored street numbers image dataset, to serve as a difficult OoD dataset. The selection is inspired by the finding in the line of works that uses a generative model for OoD detection \cite{ren2019likelihOoD, nalisnick2018deep, choi2018generative}. Those works report that a generative model of CIFAR-10 assigns higher likelihood to SVHN images, indicating a hard case for OoD detection.

\textbf{Networks and Training Details:} We use DenseNet \cite{huang2017densely}, ResNet \cite{he2016identity}, and WideResNet \cite{zagoruyko2016wide} for the classifier backbone. DenseNet has 100 layers with a growth rate of 12. It is trained with batch size 64 for 300 epochs with weight decay 0.0001. The ResNet and WideResNet-28-10 are trained with batch size 128 for 200 epochs with weight decay 0.0005. In both training, the optimizer is SGD with momentum 0.9, and the learning rate starts with 0.1 and decreases by factor 0.1 at 50\% and 75\% of the training epochs. Note that we do not apply weight decay for the weights in the $h_i(\vx)$ function of DeConf classifier since they work as the centroids for classes, and those weights are initialized with He-initialization \cite{he2015delving}. In the robustness analysis, the model may be indicated to have an extra regularization. In such case, we additional apply a dropout rate of 0.7 at the inputs for the dividend/divisor structure.

\textbf{Evaluation Metrics:} We use the two most widely adopted metrics in the OoD detection literature. The first one is the area under the receiver operating characteristic curve (AUROC), which plots the true positive rate (TPR) of in-distribution data against the false positive rate (FPR) of OoD data by varying a threshold. Thus it can be regarded as an averaged score. The second one is true negative rate at 95\% true positive rate (TNR@TPR95), which simulates an application requirement that the recall of in-distribution data should be 95\%. Having a high TNR under a high TPR is much more challenging than having a high AUROC score; thus TNR@TPR95 can discern between high-performing OoD detectors better.

\subsection{Results and Discussion} \label{sec:results}
\begin{table}
\centering
\caption{Performance of four OoD detection methods. All methods in the table have no access to OoD data during training and validation. ODIN* and Mahalanobis* are modified versions that do not need any OoD data for tuning (see Section \ref{sec:odin_maha}). The base network used in the table is DenseNet trained with CIFAR-10/100 (in-distribution data, or ID). All values are percentages averaged over three runs, and the best results are indicated in bold. Note that we only show the most common settings used in literature. The DeConf-C is selected since it shows the best robustness in our analysis, but it is not necessary to perform the best among all DeConf variants. Please see Figure \ref{fig:ab_svhn_cifar10_deconf} and Figure \ref{fig:ab_cifar_deconf} for the summary. A more comprehensive version of the table is available in Supplementary.}
\label{tbl:OoD_SOTA}
%\begin{center}
\begin{small}
%\begin{sc}
\resizebox{0.47\textwidth}{!}{
\begin{tabular}{cccc}
\toprule
ID & OoD       & AUROC  & TNR@TPR95 \\ \midrule
& & \multicolumn{2}{c}{Baseline  /  ODIN*  /  Mahalanobis*  /  DeConf-C*} \\ \midrule
\multirow{8}{*}{\rotatebox[origin=c]{90}{CIFAR-100}} &Imagenet(c)        & 79.0   /   90.5   /   92.4   /   \textbf{97.6}   &   25.3   /   56.0   /   63.5   /   \textbf{87.8} \\     
&Imagenet(r)        & 76.4   /   91.1   /   96.4   /   \textbf{98.6}   &   22.3   /   59.4   /   82.0   /   \textbf{93.3} \\     
&LSUN(c)            & 78.6   /   89.9   /   81.2   /   \textbf{95.3}   &   23.0   /   53.0   /   31.6   /   \textbf{75.0} \\     
&LSUN(r)            & 78.2   /   93.0   /   96.6   /   \textbf{98.7}   &   23.7   /   64.0   /   82.6   /   \textbf{93.8} \\     
&iSUN               & 76.8   /   91.6   /   96.5   /   \textbf{98.4}   &   21.5   /   58.4   /   81.2   /   \textbf{92.5} \\     
&SVHN               & 78.1   /   85.6   /   89.9   /   \textbf{95.9}   &   18.9   /   35.3   /   43.3   /   \textbf{77.0} \\     
&Uniform            & 65.0   /   91.4   /   \textbf{100.}   /   99.9   &   2.95   /   66.1   /   \textbf{100.}   /   \textbf{100.} \\     
&Gaussian           & 48.0   /   62.0   /   \textbf{100.}   /   99.9   &   0.06   /   33.3   /   \textbf{100.}   /   \textbf{100.} \\  \midrule
\multirow{8}{*}{\rotatebox[origin=c]{90}{CIFAR-10}}&Imagenet(c)        & 92.1   /   88.2   /   96.3   /   \textbf{98.7}   &   50.0   /   47.8   /   81.2   /   \textbf{93.4} \\     
&Imagenet(r)        & 91.5   /   90.1   /   98.2   /   \textbf{99.1}   &   47.4   /   51.9   /   90.9   /   \textbf{95.8} \\     
&LSUN(c)            & 93.0   /   91.3   /   92.2   /   \textbf{98.3}   &   51.8   /   63.5   /   64.2   /   \textbf{91.5} \\     
&LSUN(r)            & 93.9   /   92.9   /   98.2   /   \textbf{99.4}   &   56.3   /   59.2   /   91.7   /   \textbf{97.6} \\     
&iSUN               & 93.0   /   92.2   /   98.2   /   \textbf{99.4}   &   52.3   /   57.2   /   90.6   /   \textbf{97.5} \\     
&SVHN               & 88.1   /   89.6   /   98.0   /   \textbf{98.8}   &   40.5   /   48.7   /   90.6   /   \textbf{94.0} \\     
&Uniform            & 95.4   /   98.9   /   \textbf{99.9}   /   \textbf{99.9}   &   59.9   /   98.1   /   \textbf{100.}   /   \textbf{100.} \\     
&Gaussian           & 94.0   /   98.6   /   \textbf{100.}   /   99.9   &   48.8   /   92.1   /   \textbf{100.}   /   \textbf{100.} \\  \bottomrule
\end{tabular}}
%\end{sc}
\end{small}
%\end{center}
\end{table}

\begin{table}
\centering
\caption{OoD detection with OoD data versus without OoD data with CIFAR-10/100 for the in-distribution (ID) data. The values of ODIN$^{orig}$ and Maha$^{orig}$ (abbreviation of Mahalanobis) are copied from the Mahalanobis paper \cite{lee2018simple} which are tuned with OoD data. The values of ODIN*, Maha*, and DeConf-C* are copied from Table \ref{tbl:OoD_SOTA} of our paper which do not have any access to OoD data. All methods in this table use the same DenseNet for the backbone. Note that the performance with different network backbone may have a mild difference. For example, Maha$^{orig}$ performs slightly better than DeConf-C* with ResNet-34.}
\label{tbl:use_OoD}
%\begin{center}
%\begin{small}
%\begin{sc}
\resizebox{0.48\textwidth}{!}{
\begin{tabular}{cccc}
\toprule
ID & OoD       & AUROC  & TNR@TPR95 \\ \midrule
& & \multicolumn{2}{c}{ODIN$^{orig}$  / Maha$^{orig}$/  ODIN*   / Maha*  /  DeConf-C*} \\ \midrule
\multirow{3}{*}{\rotatebox[origin=c]{90}{C-100}} &Imagenet(r)        & 85.2   / 97.4  /   91.1     / 96.4   /   \textbf{98.6}   &   42.6   / 86.6  /   59.4    / 82.0   /   \textbf{93.3} \\     
&LSUN(r)            & 85.5   / 98.0    /   93.0    / 96.6   /   \textbf{98.7}   &   41.2   / 91.4  /   64.0    / 82.6   /   \textbf{93.8} \\     
&SVHN               & 93.8   / \textbf{97.2} /   85.6    / 89.9   /   95.9   &   70.6   / \textbf{82.5} /   35.3    / 43.3   /   77.0 \\ \midrule
\multirow{3}{*}{\rotatebox[origin=c]{90}{C-10}}&Imagenet(r)        & 98.5   / 98.8 /   90.1    / 98.2   /   \textbf{99.1}   &   92.4   / 95.0 /   51.9    / 90.9   /   \textbf{95.8} \\     
&LSUN(r)            & 99.2   / 99.3 /   92.9    / 98.2   /   \textbf{99.4}   &   96.2   / 97.2 /   59.2    / 91.7   /   \textbf{97.6} \\     
&SVHN               & 95.5   / 98.1 /   89.6    / 98.0   /   \textbf{98.8}   &   86.2   / 90.8 /   48.7    / 90.6   /   \textbf{94.0} \\     
\bottomrule
\end{tabular}}
%\end{sc}
%\end{small}
%\end{center}
\end{table}

\textbf{OoD benchmark performance:}
We show an overall comparison for methods that train without OoD data in Table \ref{tbl:OoD_SOTA} with 8 OoD benchmark datasets. The ODIN* and Mahalanobis* are significantly better than the baseline, while DeConf-C* still outperforms them with a significant margin. These results clearly show that learning OoD detection without OoD data is feasible, and the two methods we proposed in Sections \ref{sec:deconf} and \ref{sec:IPP} combined are very effective for this purpose.

 In Table \ref{tbl:use_OoD} we further compare our results with the original ODIN \cite{liang2017enhancing} and Mahalanobis \cite{lee2018simple} methods which are tuned on each OoD dataset. We refer to the results of both original methods reported by \cite{lee2018simple} since it uses the same backbone network, OoD datasets, and metrics to evaluate OoD detection performance. In the comparison, we find our ODIN* and Mahalanobis* perform worse than the ODIN$^{orig}$ and Mahalanobis$^{orig}$ in a major fraction of the cases. The result is not surprising because the original methods gain advantage from using OoD data. However, our DeConf-C* still outperforms the two original methods in many of the cases. The cross-setting comparison further supports the effectiveness of the proposed strategies.

\begin{figure}
  \centering
  \subfloat[CIFAR-10 classifier]{
    \includegraphics[clip, trim=0cm 8.5cm 0cm 0cm, width=0.5\textwidth]{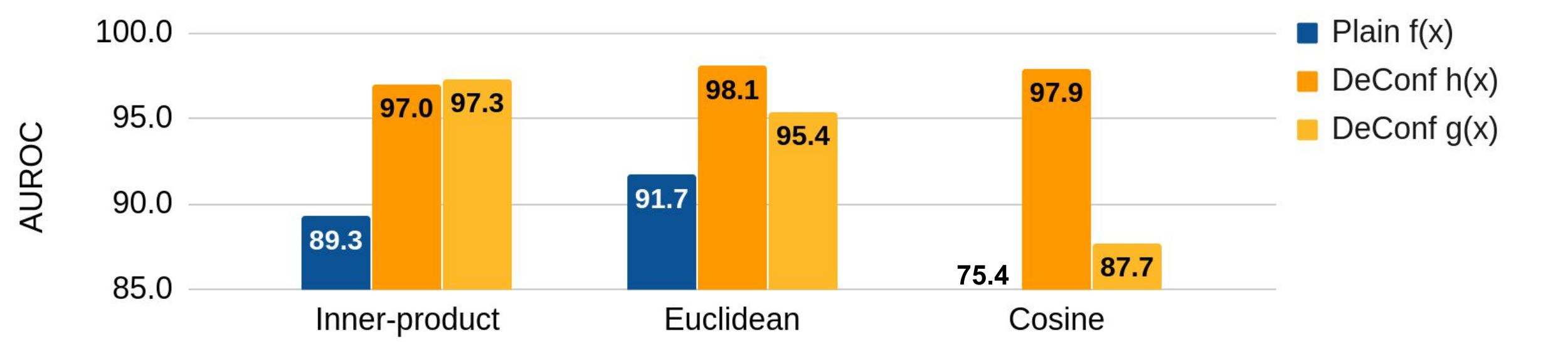}
  }
  \qquad
  \subfloat[SVHN classifier]{
    \includegraphics[clip, trim=0cm 3cm 0cm 6cm, width=0.5\textwidth]{fig/cifar10_svhn_deconf.pdf}
  }
  \caption{An ablation study with three variants in our DeConf method (Section \ref{sec:deconf}). \textit{Plain} means $g(\vx)=1$ so that the dividend/divisor structure is turned off. Each bar in the figure is averaged with 24 experiments (8 OoD datasets listed in Table \ref{tbl:OoD_SOTA} with 3 repeats. Note that we use CIFAR-10 as OoD to replace the SVHN in the case of SVHN classifier). The backbone network is Resnet-34. The \textit{plain} setting with inner-product is equivalent to a vanilla Resnet for classification. Overall, both scores from $h(x)$ and $g(x)$ are significant higher than random (AUROC=0.5) and corresponding \textit{plain} baselines. Supplementary has breakdown results.\yh{update}} \label{fig:ab_svhn_cifar10_deconf}
\end{figure}

\begin{figure}
  \centering
  \subfloat[CIFAR-100 classifier]{
    \includegraphics[clip, trim=0cm 8.5cm 0cm 0cm, width=0.5\textwidth]{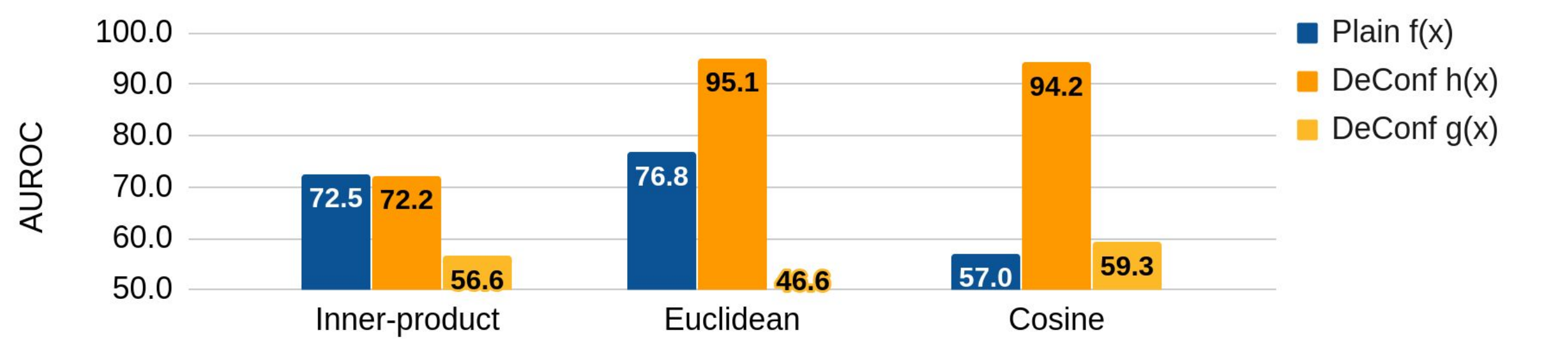}
    \label{fig:ab_cifar_deconf_a}
  }
  \qquad
  \subfloat[CIFAR-100 classifier with extra regularization (dropout 0.7)]{
    \includegraphics[clip, trim=0cm 3cm 0cm 6cm, width=0.5\textwidth]{fig/cifar100_deconf.pdf}
    \label{fig:ab_cifar_deconf_b}
  }
  \caption{An ablation study similar to Figure \ref{fig:ab_svhn_cifar10_deconf}. This figure shows the performance of DeConf-I and all $g(\vx)$ are improved by adding extra regularization.} \label{fig:ab_cifar_deconf}
\end{figure}

\begin{figure}
  \centering
  \includegraphics[clip, trim=0.5cm 6cm 0.5cm 0cm, width=0.5\textwidth]{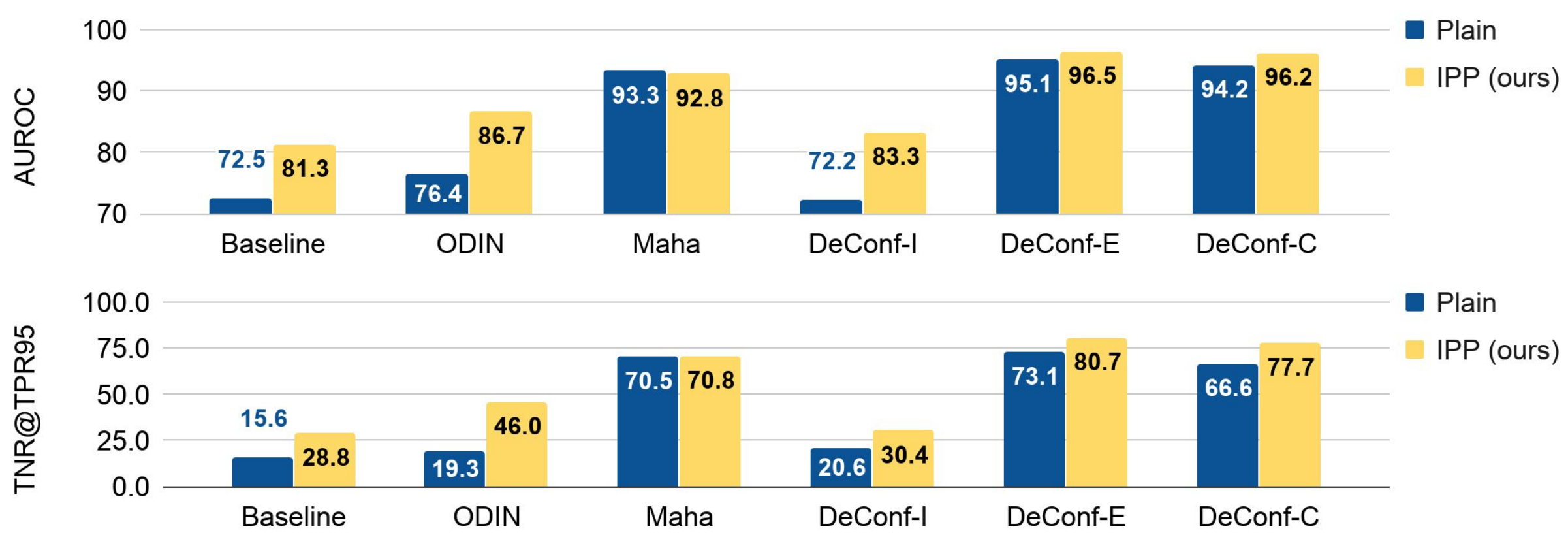} 
  \caption{The OoD detection performance of our input preprocessing (IPP) strategy, which selects the perturbation magnitude with only in-distribution data. The setting \textit{plain} means the IPP is turned off. The in-distribution data is CIFAR-100. The backbone network is Resnet-34. Each value is averaged with the results on 8 OoD datasets listed in Table \ref{tbl:OoD_SOTA}. Each method has its own scoring function $S(\vx)$ (See Section \ref{sec:odin_maha} and \ref{sec:approach}), causing IPP to perform at varied levels of performance gain. }\label{fig:ab_ipp}
\end{figure}

\textbf{Ablation Study:} We study the effect of applying DeConf and our modified input preprocessing (IPP) strategy separately. In Figure \ref{fig:ab_svhn_cifar10_deconf}, it shows that both $h_i(\vx)$ and $g(\vx)$ from all three variants (I, E, C) of the DeConf strategy help OoD detection performance with CIFAR-10 and SVHN classifiers, showing that the concept of DeConf is generally effective. However, the failure of DeConf-I and $g(x)$ with the CIFAR-100 classifier in Figure \ref{fig:ab_cifar_deconf_a} may indicate these functions have different robustness and scalability, which we will investigate in the next section. One downside of using the DeConf strategy is that the accuracy of the classifier may slightly reduce in the case of CIFAR-100 (See Table \ref{tbl:IDACC_lite}).\yh{updated} This could be a natural consequence of having an alternative term, \ie $g(\vx)$, in the model to fit the loss function. This may cause the lack of a high score for $h_i(\vx)$, instead of assigning a lower score for the data away from the high-density region of in-distribution data. We see this effect is reduced and has only a $1\%$ accuracy drop when the extra regularization (dropout rate 0.7) is applied.

In Figure \ref{fig:ab_ipp}, the results show that tuning the perturbation magnitude with only in-distribution data is an effective strategy, allowing us to reduce the required supervision for learning. The supervision here means the binary label for in/out-of-distribution.

\begin{figure}
  \centering
  \includegraphics[clip, trim=0.3cm 7.5cm 3cm 0cm, width=0.5\textwidth]{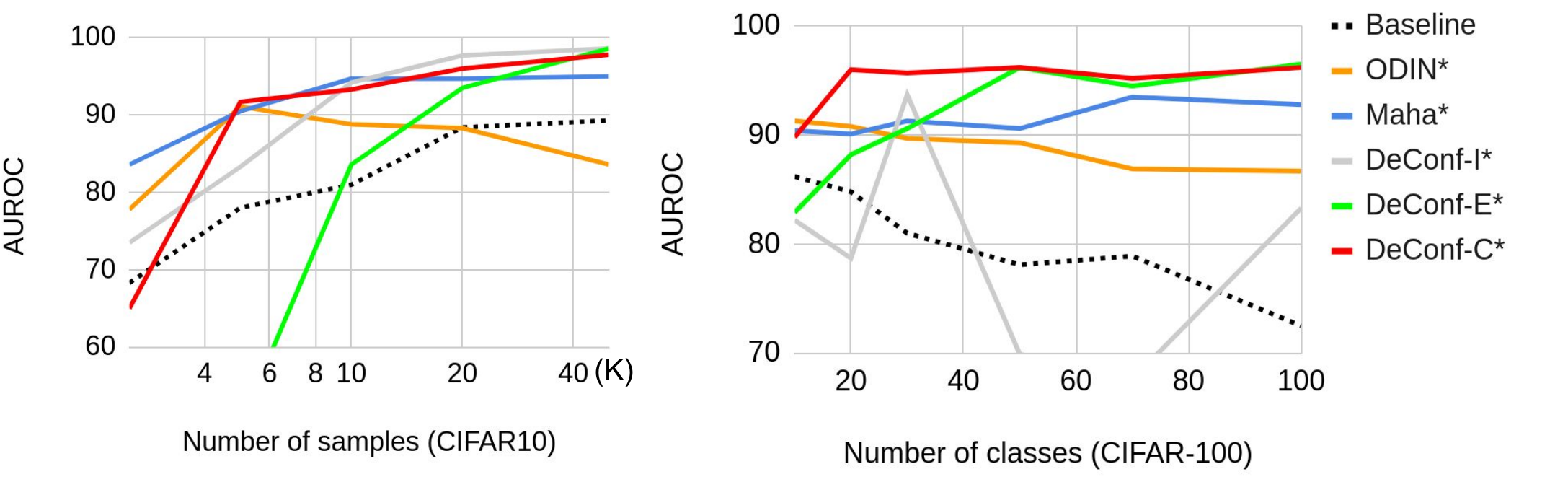} 
  \caption{Robustness analysis of 6 OoD detection methods. The left figure has classifiers trained on a varied number of samples in CIFAR-10. The right figure has classifiers trained on a varied number of classes in CIFAR-100. Each point in the line is an average of the results on 8 OoD datasets. The backbone network is Resnet-34. Please see Section \ref{sec:results} for a detailed discussion. }\label{fig:factors}
\end{figure}

\begin{figure}
  \centering
  \includegraphics[clip, trim=0.9cm 9.7cm 1cm 0cm, width=0.48\textwidth]{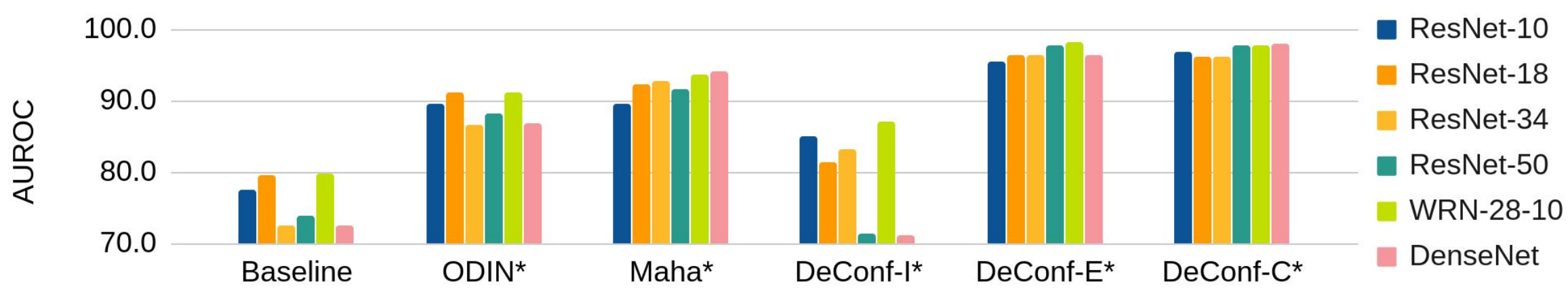} 
  \caption{Robustness analysis using different neural network backbones. The in-distribution data is CIFAR-100. Each bar is averaged with the results on 8 OoD datasets.
  }\label{fig:nets}
\end{figure}

\textbf{Robustness Study:} This study investigates when the OoD detection method will or will not work. In Figure \ref{fig:factors}, it shows that the number of in-distribution training data can largely affect the performance of the OoD detector. Mahalanobis has the lowest data requirement, but the DeConf methods generally reach a higher performance in the high data regime. In Figure \ref{fig:factors}, we also examine scalability by varying the number of classes in the in-distribution data. In this test, DeConf-E* and DeConf-C* show the best scalability. Overall, DeConf-C* is more robust than the other two DeConf variants. Lastly, Figure \ref{fig:nets} shows that high performing methods such as DeConf-E*, DeConf-C*, and Mahalanobis* are not sensitive to the type and depth of neural networks. Therefore, \textit{the number of in-distribution samples and classes are the main factors that affect OoD detection performance}. 

\begin{figure}
  \centering
  \includegraphics[clip, trim=0.4cm 6.4cm 1cm 0.4cm, width=0.48\textwidth]{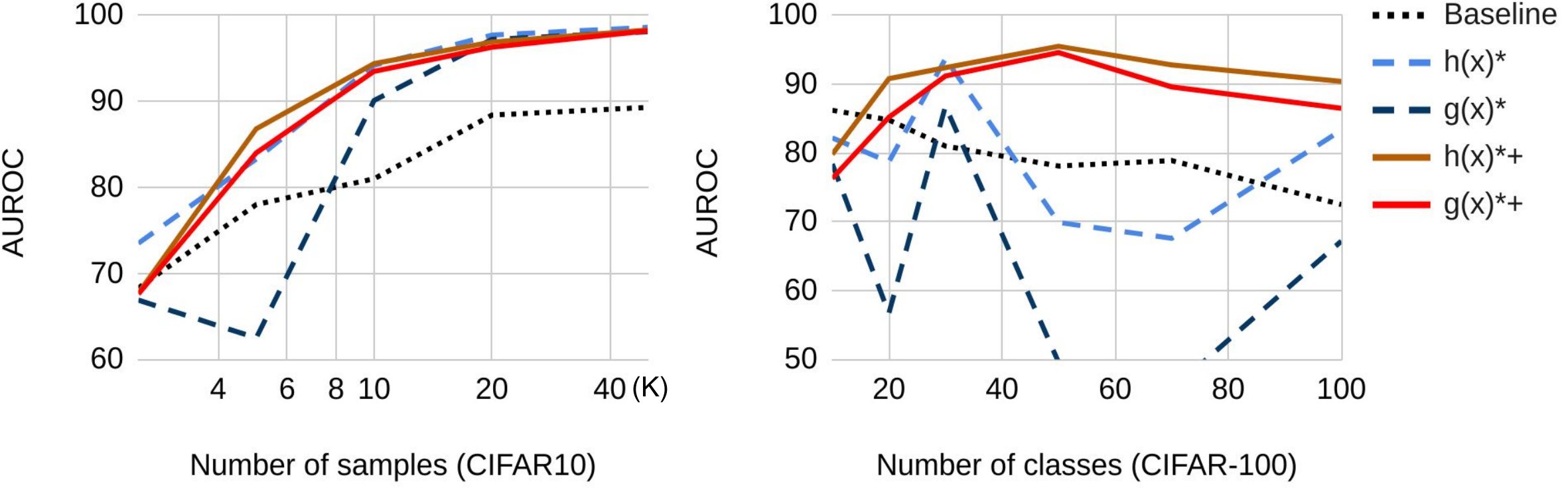} 
  \caption{Robustness analysis for $h(x)$ and $g(x)$ from DeConf-I. The + sign represents the model trained with extra regularization (dropout rate 0.7). }\label{fig:factors_dropout}
\end{figure}

\textbf{Enhancing the Robustness:} The overfitting issue may be the cause of low OoD detection performance for some of the DeConf variants and $g(x)$. In Figure \ref{fig:ab_cifar_deconf_b}, the OoD detection performance is significantly increased with DeConf-I and all $g(x)$ when extra regularization (dropout rate 0.7) is applied. Figure \ref{fig:factors_dropout} provides further analysis for DeConf-I and its $g(x)$ by varying the number of samples and classes in the training data. The performance with extra regularization is significantly better than the cases without it. Besides, the performance is also very similar between regularized $h_i(\vx)$ and $g(\vx)$, indicating that overfitting is an important issue. Lastly, we note that the DeConf-E and DeConf-C have a reduced performance with extra regularization in Figure \ref{fig:ab_cifar_deconf_b}. This outcome might be because the dropout generally harms the distance calculation between centroids and data since part of the feature is masked. The results indicate that the design of (I, E, C) might not be optimal for the problem, leaving room for future work to find a robust pair of $h_i(\vx)$ and $g(\vx)$ for the OoD detection problem.

\subsection{Semantic Shift versus Non-semantic Shift}

One interesting aspect of out-of-distribution data that has not been explored is the separation of semantic and non-semantic shift. We therefore use a larger scale image dataset, DomainNet \cite{peng2018moment}, to repeat an evaluation similar to Table \ref{tbl:OoD_SOTA}. DomainNet has high-resolution (180x180 to 640x880)\yh{updated} images in 345 classes from six different domains. There are four domains in the dataset with class labels available when the experiments were conducted.\yh{updated} They are real, sketch, infograph, and quickdraw, resulting in different types of distribution shifts.

To create subsets with semantic shift, we separate the classes into two splits. Split A has class indices from 0 to 172, while split B has 173 to 344. Our experiment uses real-A for in-distribution and has the other subsets for out-of-distribution. With the definition given in Section \ref{sec:background}, real-B has a semantic shift from real-A, while sketch-A has a non-semantic shift. Sketch-B therefore has both types of distribution shift. Figure \ref{fig:shifts} illustrates the setup. The classifier learned on real-A uses a Resnet-34 backbone. Its training setting is described in Section \ref{sec:exp_setting} except that the networks are trained for 100 epochs with initial learning rate of 0.01, and the images are center-cropped and resized to 224x224 in this experiment.

\begin{table}
\centering
\caption{The in-domain classification accuracy. The "+" means that the classifier is trained with extra regularization (dropout rate 0.7). The expanded version of this table is available in Supplementary.\yh{updated}}
\begin{small}
\label{tbl:IDACC_lite}
%\begin{center}
\centering
\resizebox{0.49\textwidth}{!}{
\begin{tabular}{cccccccc}
\toprule
Classifier       & Model   & Baseline   & DeConf-I   & DeConf-E   & DeConf-C   \\
\midrule
CIFAR-10         & DenseNet   & 95.2$\pm$0.1 & 94.9$\pm$0.1 & 95.0$\pm$0.1 & 95.0$\pm$0.1 \\
CIFAR-100        & DenseNet   & 77.0$\pm$0.2 & 75.8$\pm$0.4 & 76.4$\pm$0.1 & 75.9$\pm$0.1 \\
SVHN             & ResNet34   & 96.9$\pm$0.1 & 96.8$\pm$0.1 & 96.5$\pm$0.1 & 96.7$\pm$0.1 \\
CIFAR-10         & ResNet34   & 95.2$\pm$0.1 & 95.0$\pm$0.1 & 94.9$\pm$0.1 & 95.1$\pm$0.1 \\
CIFAR-100        & ResNet34   & 78.5$\pm$0.2 & 76.0$\pm$0.1 & 76.2$\pm$0.1 & 75.8$\pm$0.2 \\
CIFAR-100$^+$    & ResNet34   & 78.2$\pm$0.1 & 77.4$\pm$0.3 & 77.2$\pm$0.3 & 77.2$\pm$0.1 \\
\begin{tabular}{@{}c@{}}DomainNet \\ (Real-A)\end{tabular} & ResNet34  & 73.6$\pm$0.1 & 73.0$\pm$0.1   & 73.4$\pm$1.5 & 72.2$\pm$0.5 \\
\bottomrule
\end{tabular}
}
%\end{center}
\end{small}
\end{table}

\begin{table}
\centering
\caption{Performance of four OoD detection methods using DomainNet. The in-distribution is the real-A subset. Each value is averaged over three runs. The type of distribution shift presents a trend of difficulty to the OoD detection problem: Semantic shift (S) $>$ Non-semantic shift (NS) $>$ Semantic $+$ Non-semantic shift.}
\label{tbl:domainnet}
%\begin{center}
\begin{small}
%\begin{sc}
\resizebox{0.49\textwidth}{!}{
\begin{tabular}{ccccc}
\toprule
OOD       & \multicolumn{2}{c}{Shift} & AUROC  & TNR@TPR95 \\ \midrule
& S & NS & \multicolumn{2}{c}{Baseline  /  ODIN*  /  Maha*  /  DeConf-C*} \\ \midrule
real-B           & \checkmark&            & \textbf{75.1}   /   69.9   /   53.6   /   69.8     & 15.3   /   \textbf{15.4}   /   5.09   /   14.0  \\  
sketch-A         &           & \checkmark & 75.5   /   80.7   /   59.5   /   \textbf{84.5}     & 20.1   /   31.2   /   7.30   /   \textbf{37.5}  \\      
sketch-B         & \checkmark& \checkmark & 81.8   /   85.7   /   60.4   /   \textbf{89.1}     & 25.2   /   36.8   /   7.55   /   \textbf{44.1}  \\  
infograph-A      &           & \checkmark & 79.6   /   82.7   /   81.5   /   \textbf{89.0}     & 23.5   /   27.8   /   21.6   /   \textbf{45.4}  \\  
infograph-B      & \checkmark& \checkmark & 82.1   /   85.3   /   80.9   /   \textbf{90.9}     & 24.8   /   31.7   /   21.9   /   \textbf{49.6}  \\      
quickdraw-A      &           & \checkmark & 78.8   /   96.4   /   67.4   /   \textbf{96.9}     & 21.1   /   79.9   /   3.38   /   \textbf{83.1}  \\      
quickdraw-B      & \checkmark& \checkmark & 80.5   /   96.9   /   66.1   /   \textbf{97.4}     & 22.1   /   83.6   /   2.38   /   \textbf{86.6}  \\      
Uniform          & \checkmark& \checkmark & 54.7   /   75.6   /   \textbf{99.8}   /   99.3     & 1.65   /   5.37   /   \textbf{100.}   /   \textbf{100.}  \\      
Gaussian         & \checkmark& \checkmark & 71.3   /   95.5   /   \textbf{99.9}   /   99.4     & 0.64   /   46.9   /   \textbf{100.}   /   \textbf{100.}  \\      
\bottomrule
\end{tabular}}
%\end{sc}
\end{small}
%\end{center}
\end{table}

The results in Table \ref{tbl:domainnet} reveal two interesting trends. The first one is that the OoD datasets with both types of distribution shifts are easier to detect, followed by non-semantic shift. The semantic shift turns out to be the hardest one to detect. The second observation is the failure of Mahalanobis*. In most cases it is even worse than Baseline, except detecting random noise. In contrast, ODIN* has performance gain in most of the cases, but has less gain with random noise. Our DeConf-C* still performs the best, showing that its robustness and scalability is capable of handling a more realistic problem setting, although there is still large room for improvement.

\section{Conclusion}

In this paper, we propose two strategies, the decomposed confidence and the modified input preprocessing. These two simple modifications to ODIN lead to a significant change in the paradigm, which does not need OoD data for tuning the method. Our comprehensive analysis shows that our strategies are effective and even better in several cases than the methods tuned for each OoD dataset. Our further analysis using a larger scale image dataset shows that the data with only semantic shift is harder to detect, pointing out a challenge for future works to address.

{\small
\bibliographystyle{ieee_fullname}
\bibliography{main}

\begin{thebibliography}{10}\itemsep=-1pt

\bibitem{andrews2016transfer}
J Andrews, Thomas Tanay, Edward~J Morton, and Lewis~D Griffin.
\newblock Transfer representation-learning for anomaly detection.
\newblock In {\em JMLR}, 2016.

\bibitem{bendale2015towards}
Abhijit Bendale and Terrance Boult.
\newblock Towards open world recognition.
\newblock In {\em Proceedings of the IEEE Conference on Computer Vision and
  Pattern Recognition}, pages 1893--1902, 2015.

\bibitem{bendale2016towards}
Abhijit Bendale and Terrance~E Boult.
\newblock Towards open set deep networks.
\newblock In {\em Proceedings of the IEEE conference on computer vision and
  pattern recognition}, pages 1563--1572, 2016.

\bibitem{choi2018generative}
Hyunsun Choi and Eric Jang.
\newblock Generative ensembles for robust anomaly detection.
\newblock {\em arXiv preprint arXiv:1810.01392}, 2018.

\bibitem{CortesRejection}
Corinna Cortes, Giulia DeSalvo, and Mehryar Mohri.
\newblock Learning with rejection.
\newblock In {\em International Conference on Algorithmic Learning Theory},
  pages 67--82. Springer, 2016.

\bibitem{Dhamija18}
Akshay~Raj Dhamija, Manuel G\"{u}nther, and Terrance Boult.
\newblock Reducing network agnostophobia.
\newblock {\em Advances in Neural Information Processing Systems}, 2018.

\bibitem{gal2016dropout}
Yarin Gal and Zoubin Ghahramani.
\newblock Dropout as a bayesian approximation: Representing model uncertainty
  in deep learning.
\newblock In {\em international conference on machine learning}, pages
  1050--1059, 2016.

\bibitem{geifman2017selective}
Yonatan Geifman and Ran El-Yaniv.
\newblock Selective classification for deep neural networks.
\newblock In {\em Advances in neural information processing systems}, pages
  4878--4887, 2017.

\bibitem{guo2017calibration}
Chuan Guo, Geoff Pleiss, Yu Sun, and Kilian~Q Weinberger.
\newblock On calibration of modern neural networks.
\newblock In {\em Proceedings of the 34th International Conference on Machine
  Learning}, 2017.

\bibitem{he2015delving}
Kaiming He, Xiangyu Zhang, Shaoqing Ren, and Jian Sun.
\newblock Delving deep into rectifiers: Surpassing human-level performance on
  imagenet classification.
\newblock In {\em Proceedings of the IEEE international conference on computer
  vision}, pages 1026--1034, 2015.

\bibitem{he2016identity}
Kaiming He, Xiangyu Zhang, Shaoqing Ren, and Jian Sun.
\newblock Identity mappings in deep residual networks.
\newblock In {\em European conference on computer vision}, pages 630--645.
  Springer, 2016.

\bibitem{hellman1970nearest}
Martin~E Hellman.
\newblock The nearest neighbor classification rule with a reject option.
\newblock {\em IEEE Transactions on Systems Science and Cybernetics},
  6(3):179--185, 1970.

\bibitem{hendrycks2016baseline}
Dan Hendrycks and Kevin Gimpel.
\newblock A baseline for detecting misclassified and out-of-distribution
  examples in neural networks.
\newblock {\em International Conference on Learning Representations (ICLR)},
  2017.

\bibitem{hendrycks2018deep}
Dan Hendrycks, Mantas Mazeika, and Thomas~G Dietterich.
\newblock Deep anomaly detection with outlier exposure.
\newblock {\em International Conference on Learning Representations (ICLR)},
  2019.

\bibitem{hinton2015distilling}
Geoffrey Hinton, Oriol Vinyals, and Jeff Dean.
\newblock Distilling the knowledge in a neural network.
\newblock {\em arXiv preprint arXiv:1503.02531}, 2015.

\bibitem{huang2017densely}
Gao Huang, Zhuang Liu, Laurens Van Der~Maaten, and Kilian~Q Weinberger.
\newblock Densely connected convolutional networks.
\newblock In {\em Proceedings of the IEEE conference on computer vision and
  pattern recognition}, pages 4700--4708, 2017.

\bibitem{krizhevsky2009learning}
Alex Krizhevsky et~al.
\newblock Learning multiple layers of features from tiny images.
\newblock Technical report, Citeseer, 2009.

\bibitem{lakshminarayanan2017simple}
Balaji Lakshminarayanan, Alexander Pritzel, and Charles Blundell.
\newblock Simple and scalable predictive uncertainty estimation using deep
  ensembles.
\newblock In {\em Advances in Neural Information Processing Systems}, pages
  6402--6413, 2017.

\bibitem{lee2017training}
Kimin Lee, Honglak Lee, Kibok Lee, and Jinwoo Shin.
\newblock Training confidence-calibrated classifiers for detecting
  out-of-distribution samples.
\newblock {\em International Conference on Learning Representations (ICLR)},
  2018.

\bibitem{lee2018simple}
Kimin Lee, Kibok Lee, Honglak Lee, and Jinwoo Shin.
\newblock A simple unified framework for detecting out-of-distribution samples
  and adversarial attacks.
\newblock In {\em Advances in Neural Information Processing Systems}, pages
  7167--7177, 2018.

\bibitem{liang2017enhancing}
Shiyu Liang, Yixuan Li, and R Srikant.
\newblock Enhancing the reliability of out-of-distribution image detection in
  neural networks.
\newblock {\em arXiv preprint arXiv:1706.02690}, 2017.

\bibitem{MalininNIPS2018}
Andrey Malinin and Mark Gales.
\newblock Predictive uncertainty estimation via prior networks.
\newblock In S. Bengio, H. Wallach, H. Larochelle, K. Grauman, N. Cesa-Bianchi,
  and R. Garnett, editors, {\em Advances in Neural Information Processing
  Systems 31}, pages 7047--7058. Curran Associates, Inc., 2018.

\bibitem{malinin2019reverse}
Andrey Malinin and Mark Gales.
\newblock Reverse kl-divergence training of prior networks: Improved
  uncertainty and adversarial robustness.
\newblock In {\em Advances in Neural Information Processing Systems}, pages
  14520--14531, 2019.

\bibitem{malinin2019ensemble}
Andrey Malinin, Bruno Mlodozeniec, and Mark Gales.
\newblock Ensemble distribution distillation.
\newblock In {\em International Conference on Learning Representations}, 2019.

\bibitem{masana2018metric}
Marc Masana, Idoia Ruiz, Joan Serrat, Joost van~de Weijer, and Antonio~M Lopez.
\newblock Metric learning for novelty and anomaly detection.
\newblock {\em arXiv preprint arXiv:1808.05492}, 2018.

\bibitem{nalisnick2018deep}
Eric Nalisnick, Akihiro Matsukawa, Yee~Whye Teh, Dilan Gorur, and Balaji
  Lakshminarayanan.
\newblock Do deep generative models know what they don't know?
\newblock {\em arXiv preprint arXiv:1810.09136}, 2018.

\bibitem{netzer2011reading}
Yuval Netzer, Tao Wang, Adam Coates, Alessandro Bissacco, Bo Wu, and Andrew~Y.
  Ng.
\newblock Reading digits in natural images with unsupervised feature learning.
\newblock In {\em NIPS Workshop on Deep Learning and Unsupervised Feature
  Learning}, 2011.

\bibitem{neumann2018relaxed}
Lukas Neumann, Andrew Zisserman, and Andrea Vedaldi.
\newblock Relaxed softmax: Efficient confidence auto-calibration for safe
  pedestrian detection.
\newblock {\em NIPS MLITS Workshop}, 2018.

\bibitem{nguyen2015deep}
Anh Nguyen, Jason Yosinski, and Jeff Clune.
\newblock Deep neural networks are easily fooled: High confidence predictions
  for unrecognizable images.
\newblock In {\em Proceedings of the IEEE conference on computer vision and
  pattern recognition}, pages 427--436, 2015.

\bibitem{patel2015visual}
Vishal~M Patel, Raghuraman Gopalan, Ruonan Li, and Rama Chellappa.
\newblock Visual domain adaptation: A survey of recent advances.
\newblock {\em IEEE signal processing magazine}, 32(3):53--69, 2015.

\bibitem{peng2018moment}
Xingchao Peng, Qinxun Bai, Xide Xia, Zijun Huang, Kate Saenko, and Bo Wang.
\newblock Moment matching for multi-source domain adaptation.
\newblock {\em arXiv preprint arXiv:1812.01754}, 2018.

\bibitem{ren2019likelihOoD}
Jie Ren, Peter~J Liu, Emily Fertig, Jasper Snoek, Ryan Poplin, Mark~A DePristo,
  Joshua~V Dillon, and Balaji Lakshminarayanan.
\newblock Likelihood ratios for out-of-distribution detection.
\newblock {\em arXiv preprint arXiv:1906.02845}, 2019.

\bibitem{salimans2017pixelcnn}
Tim Salimans, Andrej Karpathy, Xi Chen, and Diederik~P Kingma.
\newblock Pixelcnn++: Improving the pixelcnn with discretized logistic mixture
  likelihood and other modifications.
\newblock {\em International Conference on Learning Representations (ICLR)},
  2017.

\bibitem{Shafaei2019}
Alireza Shafaei, Mark Schmidt, and James Little.
\newblock A less biased evaluation of ood sample detectors.
\newblock In {\em Proceedings of the British Machine Vision Conference (BMVC)},
  2019.

\bibitem{shalev2018out}
Gabi Shalev, Yossi Adi, and Joseph Keshet.
\newblock Out-of-distribution detection using multiple semantic label
  representations.
\newblock In {\em Advances in Neural Information Processing Systems}, pages
  7375--7385, 2018.

\bibitem{szegedy2013intriguing}
Christian Szegedy, Wojciech Zaremba, Ilya Sutskever, Joan Bruna, Dumitru Erhan,
  Ian Goodfellow, and Rob Fergus.
\newblock Intriguing properties of neural networks.
\newblock {\em arXiv preprint arXiv:1312.6199}, 2013.

\bibitem{OOD_temp_cosine}
Engkarat Techapanurak and Takayuki Okatani.
\newblock Hyperparameter-free out-of-distribution detection using softmax of
  scaled cosine similarity.
\newblock {\em arXiv preprint:1905.10628}, 2019.

\bibitem{vyas2018out}
Apoorv Vyas, Nataraj Jammalamadaka, Xia Zhu, Dipankar Das, Bharat Kaul, and
  Theodore~L Willke.
\newblock Out-of-distribution detection using an ensemble of self supervised
  leave-out classifiers.
\newblock In {\em Proceedings of the European Conference on Computer Vision
  (ECCV)}, pages 550--564, 2018.

\bibitem{zagoruyko2016wide}
Sergey Zagoruyko and Nikos Komodakis.
\newblock Wide residual networks.
\newblock {\em arXiv preprint arXiv:1605.07146}, 2016.

\bibitem{zhang2019adacos}
Xiao Zhang, Rui Zhao, Yu Qiao, Xiaogang Wang, and Hongsheng Li.
\newblock Adacos: Adaptively scaling cosine logits for effectively learning
  deep face representations.
\newblock In {\em Proceedings of the IEEE Conference on Computer Vision and
  Pattern Recognition}, pages 10823--10832, 2019.

\end{thebibliography}
}

\iftrue
%\newpage
\clearpage
\appendix
\onecolumn
\section*{Supplementary}
\addcontentsline{toc}{section}{Supplementary}
\begin{figure}[h]
  \centering
  \includegraphics[clip, trim=0.5cm 0cm 1.5cm 0cm, width=\textwidth]{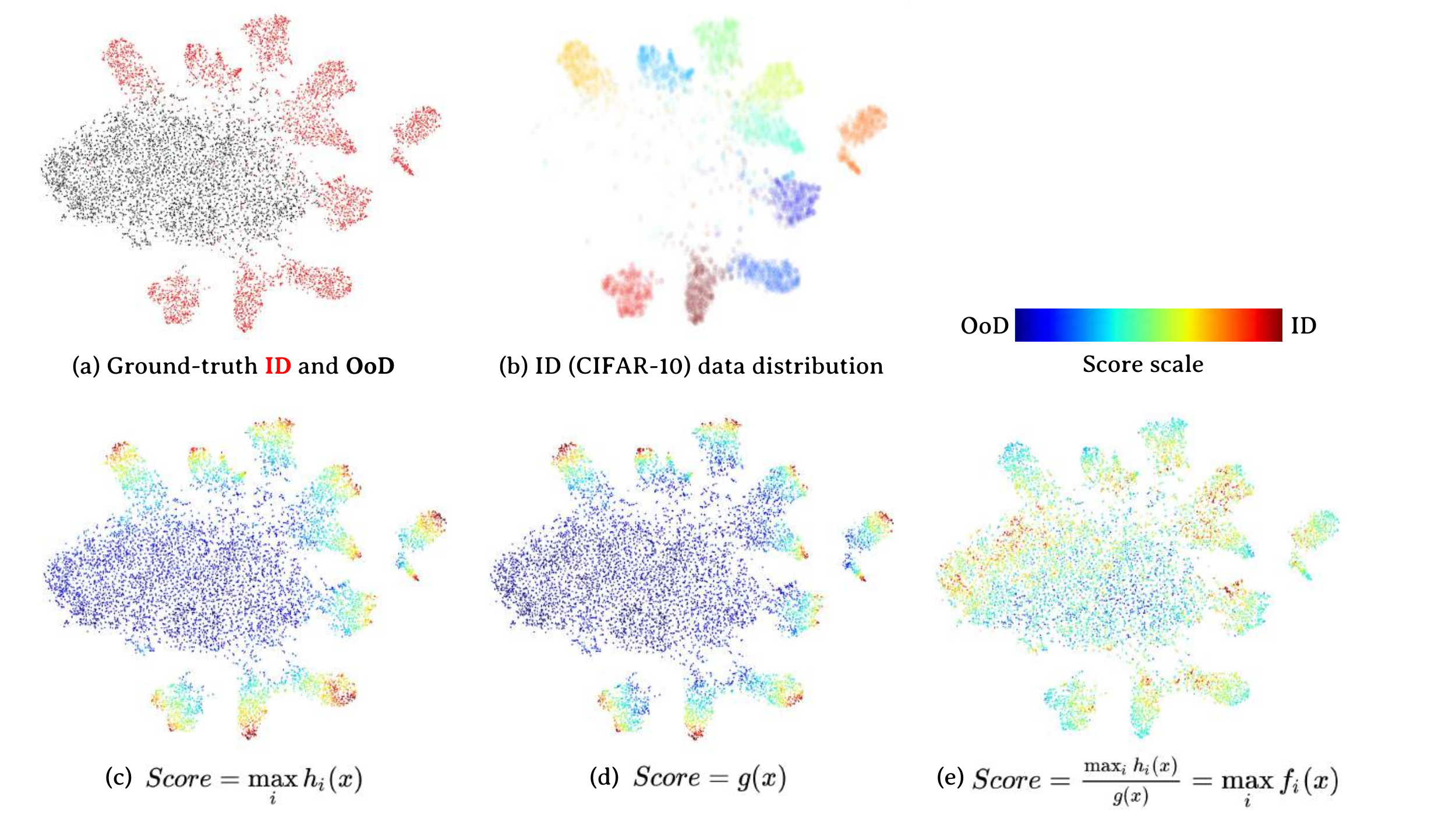} 
  \caption{Visualization of the score distribution. The data are visualized with t-SNE using the features from the penultimate layer of the neural networks. The results are from DeConf-I with ResNet-34. The figure (a) visualizes the ground-truth in-distribution (ID, red, CIFAR-10) and out-of-distribution (OoD, black, Imagenet-resized) data. The colors in (b) represent different classes of CIFAR-10. The scores are obtained from (c) $h$ function, (d) $g$ function, or (e) the logits, and the scores are linearly re-scaled to between zero and one for visualization. The figure presents two phenomena. The first is that the OoD data in (e) have high scores. It is related to the overconfident effect discussed with equation 4. The second phenomenon is that high-score data in (c) and (d) are more significantly clustered in each class of CIFAR-10. It shows a tendency that the in-distribution data in high-density regions have higher scores than those in low-density regions (close to OoD data). This phenomenon is related to the discussion at the end of section 3.1.\yh{updated}}\label{fig:deconf_viz}
\end{figure}

%\begin{sidewaystable}
\begin{table}[htb]
\caption{Performance of six OOD detection methods on 8 benchmark datasets. This is a full version of Table 1 in the main paper, which uses DenseNet for the backbone networks. All DeConf results are from the $h(x)$ branch. The value in parentheses is the standard deviation. }
\label{tbl:OoD_full}
\begin{center}
\begin{small}
%\begin{sc}
\resizebox{0.6\textwidth}{!}{
\begin{tabular}{ccc}
\toprule
ID & OOD       & AUROC \\ \midrule
& & Baseline  /  ODIN*  /  Maha*  /  DeConf-I* / DeConf-E* / DeConf-C* \\ \midrule
\multirow{8}{*}{\rotatebox[origin=c]{90}{CIFAR-100}} 
&Imagenet(c)        & 79.0(2.2) /90.5(1.1) /92.4(0.3) /84.4(2.3) /95.1(0.5) /97.6(0.2)\\
&Imagenet(r)        & 76.4(3.2) /91.1(1.3) /96.4(0.2) /81.2(3.6) /97.4(0.3) /98.6(0.2)\\
&LSUN(c)            & 78.6(1.1) /89.9(0.5) /81.2(0.6) /91.7(0.3) /90.1(0.3) /95.3(0.4)\\
&LSUN(r)            & 78.2(2.4) /93.0(0.8) /96.6(0.2) /84.1(2.1) /97.8(0.2) /98.7(0.0)\\
&iSUN               & 76.8(2.7) /91.6(1.1) /96.5(0.2) /82.1(2.9) /97.4(0.2) /98.4(0.0)\\
&SVHN               & 78.1(3.5) /85.6(0.0) /89.9(0.2) /89.7(0.4) /94.0(0.6) /95.9(0.7)\\
&Uniform            & 65.0(22.) /91.4(10.) /100.(0.0) /48.5(16.) /99.9(0.0) /99.9(0.0)\\
&Gaussian           & 48.0(28.) /62.0(38.) /100.(0.0) /6.79(4.9) /99.9(0.0) /99.9(0.0)\\
\midrule
\multirow{8}{*}{\rotatebox[origin=c]{90}{CIFAR-10}}
&Imagenet(c)        & 92.1(1.0) /88.2(4.2) /96.3(0.1) /98.2(0.0) /98.0(0.2) /98.7(0.1)\\
&Imagenet(r)        & 91.5(1.4) /90.1(4.1) /98.2(0.0) /98.4(0.0) /98.2(0.2) /99.1(0.1)\\
&LSUN(c)            & 93.0(0.5) /91.3(2.0) /92.2(0.4) /98.4(0.0) /98.6(0.2) /98.3(0.2)\\
&LSUN(r)            & 93.9(0.4) /92.9(2.9) /98.2(0.0) /98.6(0.0) /98.8(0.0) /99.4(0.1)\\
&iSUN               & 93.0(0.7) /92.2(3.4) /98.2(0.0) /98.6(0.0) /98.8(0.0) /99.4(0.0)\\
&SVHN               & 88.1(4.8) /89.6(0.3) /98.0(0.3) /98.2(0.2) /98.4(0.6) /98.8(0.1)\\
&Uniform            & 95.4(0.7) /98.9(0.7) /99.9(0.0) /99.2(0.5) /99.9(0.0) /99.9(0.0)\\
&Gaussian           & 94.0(2.9) /98.6(1.7) /100.(0.0) /99.1(0.3) /99.9(0.0) /99.9(0.0)\\
\midrule
\midrule
ID & OOD       & TNR@TPR95 \\ \midrule
& & Baseline  /  ODIN*  /  Maha*  /  DeConf-I* / DeConf-E* / DeConf-C* \\ \midrule
\multirow{8}{*}{\rotatebox[origin=c]{90}{CIFAR-100}} 
&Imagenet(c)        &25.3(2.8) /56.0(3.1) /63.5(2.1) /31.0(3.4) /74.6(2.8) /87.8(1.7)\\
&Imagenet(r)        &22.3(3.1) /59.4(3.7) /82.0(1.6) /21.4(4.0) /87.6(1.7) /93.3(1.2)\\
&LSUN(c)            &23.0(2.2) /53.0(1.0) /31.6(1.3) /59.6(1.9) /51.0(1.0) /75.0(1.9)\\
&LSUN(r)            &23.7(2.5) /64.0(3.0) /82.6(1.8) /21.1(3.3) /89.8(1.5) /93.8(0.3)\\
&iSUN               &21.5(2.8) /58.4(4.1) /81.2(1.4) /17.6(3.3) /87.3(1.2) /92.5(0.2)\\
&SVHN               &18.9(4.9) /35.3(2.9) /43.3(2.7) /52.0(0.6) /67.1(3.4) /77.0(5.0)\\
&Uniform            &2.95(4.1) /66.1(46.) /100.(0.0) /0.0(0.0)  /100.(0.0) /100.(0.0)\\
&Gaussian           &0.06(0.0) /33.3(47.) /100.(0.0) /0.0(0.0)  /100.(0.0) /100.(0.0)\\
\midrule
\multirow{8}{*}{\rotatebox[origin=c]{90}{CIFAR-10}}
&Imagenet(c)         &50.0(2.8) /47.8(15.) /81.2(0.8) /92.0(0.2) /90.1(1.5) /93.4(1.2)\\
&Imagenet(r)         &47.4(4.4) /51.9(16.) /90.9(0.5) /93.6(0.2) /91.7(1.6) /95.8(0.9)\\
&LSUN(c)             &51.8(3.1) /63.5(7.8) /64.2(0.6) /92.5(0.4) /93.3(1.5) /91.5(1.2)\\
&LSUN(r)             &56.3(3.6) /59.2(18.) /91.7(0.3) /94.9(0.2) /95.7(0.1) /97.6(0.5)\\
&iSUN                &52.3(3.6) /57.2(18.) /90.6(0.7) /94.6(0.3) /95.4(0.2) /97.5(0.3)\\
&SVHN                &40.5(6.9) /48.7(3.2) /90.6(1.7) /91.4(1.1) /92.1(3.4) /94.0(0.6)\\
&Uniform             &59.9(12.) /98.1(2.6) /100.(0.0) /99.9(0.0) /100.(0.0) /100.(0.0)\\
&Gaussian            &48.8(26.) /92.1(11.) /100.(0.0) /99.9(0.0) /100.(0.0) /100.(0.0)\\
\bottomrule
\end{tabular}}
%\end{sc}
\end{small}
\end{center}
\end{table}
%\end{sidewaystable}
%\begin{sidewaystable}
\begin{table}[htb]
\caption{Performance of six OOD detection methods on 8 benchmark datasets. The experiment here is the same as Table 1 but use Resnet-34 for the backbone network. All DeConf results are from the $h(x)$ branch. The value in parentheses is the standard deviation.}
%\label{tbl:OOD_SOTA}
\begin{center}
\begin{small}
%\begin{sc}
\resizebox{0.6\textwidth}{!}{
\begin{tabular}{ccc}
\toprule
ID & OOD       & AUROC \\ \midrule
& & Baseline  /  ODIN*  /  Maha*  /  DeConf-I* / DeConf-E* / DeConf-C* \\ \midrule
\multirow{8}{*}{\rotatebox[origin=c]{90}{CIFAR-100}} 
&Imagenet(c)       &  78.9(0.1) /84.8(0.6) /93.4(0.3) /88.2(0.6) /95.2(0.6) /95.3(0.6)\\
&Imagenet(r)       &  75.1(0.8) /85.7(0.2) /96.3(0.1) /84.6(1.0) /97.0(0.4) /95.9(0.7)\\
&LSUN(c)           &  78.8(0.6) /80.3(1.3) /79.8(0.3) /93.8(0.3) /92.6(0.2) /93.8(0.3)\\
&LSUN(r)           &  76.2(1.4) /86.6(0.8) /96.3(0.2) /85.9(1.8) /97.0(0.7) /96.1(0.5)\\
&iSUN              &  75.2(1.4) /85.9(0.8) /95.8(0.2) /84.7(1.4) /96.6(0.6) /95.7(0.5)\\
&SVHN              &  75.1(2.5) /80.2(2.0) /80.9(1.1) /89.2(2.6) /93.8(0.8) /93.2(1.1)\\
&Uniform           &  69.0(13.) /96.7(2.5) /100.(0.0) /79.3(8.3) /99.9(0.0) /99.9(0.0)\\
&Gaussian          &  51.5(1.8) /93.7(1.7) /99.9(0.0) /60.8(23.) /99.9(0.0) /99.9(0.0)\\
\midrule
\multirow{8}{*}{\rotatebox[origin=c]{90}{CIFAR-10}}
&Imagenet(c)       &  90.0(0.9) /81.2(2.4) /94.2(0.1) /98.2(0.2) /98.2(0.1) /96.0(0.2)\\
&Imagenet(r)       &  87.3(1.3) /81.1(2.9) /96.5(0.1) /98.1(0.3) /98.1(0.3) /96.1(0.5)\\
&LSUN(c)           &  92.0(1.7) /77.9(4.6) /87.7(0.2) /98.8(0.1) /98.5(0.0) /97.2(0.1)\\
&LSUN(r)           &  91.6(1.2) /88.5(2.0) /97.2(0.1) /98.9(0.2) /99.0(0.1) /98.0(0.1)\\
&iSUN              &  90.1(1.4) /86.1(2.5) /96.5(0.2) /98.8(0.2) /98.9(0.1) /97.6(0.1)\\
&SVHN              &  87.7(2.4) /63.9(4.3) /87.8(1.6) /96.8(0.4) /96.1(1.4) /97.8(0.3)\\
&Uniform           &  85.9(10.) /93.3(4.5) /99.9(0.0) /99.6(0.1) /99.9(0.0) /99.9(0.0)\\
&Gaussian          &  89.9(10.) /97.1(2.0) /99.9(0.0) /99.7(0.0) /99.9(0.0) /99.9(0.0)\\
\midrule
\midrule
ID & OOD       & TNR@TPR95 \\ \midrule
& & Baseline  /  ODIN*  /  Maha*  /  DeConf-I* / DeConf-E* / DeConf-C* \\ \midrule
\multirow{8}{*}{\rotatebox[origin=c]{90}{CIFAR-100}} 
&Imagenet(c)       &24.1(0.6) /44.0(2.2) /68.2(1.4) /42.6(2.7) /73.4(3.7) /72.6(3.7)\\
&Imagenet(r)       &19.4(0.1) /45.5(1.4) /82.6(0.8) /30.4(3.0) /84.3(2.7) /76.5(3.8)\\
&LSUN(c)           &21.9(0.4) /34.8(2.4) /27.7(1.4) /66.1(2.2) /59.7(0.7) /65.7(2.3)\\
&LSUN(r)           &19.8(1.6) /48.2(3.0) /81.8(1.4) /29.4(5.2) /84.6(4.0) /76.8(3.3)\\
&iSUN              &17.7(0.5) /45.3(2.8) /80.4(0.8) /27.1(4.3) /83.0(3.1) /75.3(3.3)\\
&SVHN              &16.6(1.5) /27.5(5.0) /25.7(2.6) /43.7(10.) /60.8(5.3) /55.1(7.1)\\
&Uniform           &5.63(7.0) /76.4(27.) /100.(0.0) /4.11(5.8) /100.(0.0) /100.(0.0)\\
&Gaussian          &0.0(0.0)  /46.6(20.) /100.(0.0) /0.06(0.0) /100.(0.0) /100.(0.0)\\
\midrule
\multirow{8}{*}{\rotatebox[origin=c]{90}{CIFAR-10}}
&Imagenet(c)       &54.6(2.6) /53.7(3.1) /74.6(0.6) /90.8(1.5) /91.1(0.9) /81.1(1.7)\\
&Imagenet(r)       &48.3(3.2) /53.1(4.3) /85.1(0.6) /90.5(1.8) /90.8(1.8) /81.4(2.4)\\
&LSUN(c)           &59.9(4.7) /50.9(6.1) /53.6(1.0) /93.9(0.5) /92.4(0.5) /87.3(1.0)\\
&LSUN(r)           &57.5(4.4) /68.1(4.2) /87.4(0.8) /95.8(1.0) /96.0(0.7) /90.9(0.9)\\
&iSUN              &53.7(3.8) /62.8(5.0) /84.6(0.9) /95.1(1.0) /95.3(0.5) /88.8(1.1)\\
&SVHN              &44.5(8.1) /29.7(6.2) /46.2(4.8) /84.5(2.5) /78.8(7.6) /89.5(2.1)\\
&Uniform           &27.9(20.) /74.5(20.) /100.(0.0) /100.(0.0) /100.(0.0) /100.(0.0)\\
&Gaussian          &52.7(40.) /87.1(9.3) /100.(0.0) /100.(0.0) /100.(0.0) /100.(0.0)\\
\bottomrule
\end{tabular}}
%\end{sc}
\end{small}
\end{center}
\end{table}
%\end{sidewaystable}
%\begin{sidewaystable}
\begin{table}[htb]
\centering
\caption{The AUROC of individual experimental setting in Figures 3 
%\ref{fig:ab_svhn_cifar10_deconf} 
and 4
%\ref{fig:ab_cifar_deconf}
. The experiments do not use input preprocessing. All values are percentages averaged over three runs, and the value in parentheses is the standard deviation. The "+" means that the classifier is trained with extra regularization (dropout rate 0.7).\yh{updated}}
%\label{tbl:OOD_SOTA}
%\begin{center}
\begin{small}
%\begin{sc}
\resizebox{1.02\textwidth}{!}{
\begin{tabular}{ccccccccccc}
\toprule
ID & OoD & Plain-I  &  DeConf-I-$h(x)$ & DeConf-I-$g(x)$ & Plain-E & DeConf-E-$h(x)$ & DeConf-E-$g(x)$ & Plain-C & DeConf-C-$h(x)$ & DeConf-C-$g(x)$ \\ \midrule
\multirow{9}{*}{\rotatebox[origin=c]{90}{SVHN}} 
&Imagenet(c)         &92.8(1.0) &98.7(0.1) &98.4(0.0) &92.9(0.8) &96.2(0.4) &97.2(0.7) &60.3(1.4) &93.9(0.4) &91.3(6.0)\\ 
&Imagenet(r)         &92.4(0.8) &98.7(0.1) &98.4(0.0) &93.0(0.6) &96.1(0.4) &97.0(0.9) &63.9(1.5) &93.5(0.7) &90.8(6.0)\\ 
&LSUN(c)             &91.0(0.6) &98.0(0.2) &97.5(0.5) &92.0(0.6) &95.0(0.1) &95.5(0.4) &51.9(0.3) &93.1(0.3) &92.4(5.8)\\ 
&LSUN(r)             &91.3(1.0) &98.4(0.2) &98.3(0.3) &92.1(0.7) &95.5(0.7) &96.7(1.1) &61.4(1.6) &92.0(0.7) &90.6(5.8)\\ 
&iSUN                &91.5(0.9) &98.6(0.1) &98.5(0.2) &92.4(0.8) &95.8(0.6) &96.7(1.1) &60.9(1.6) &93.1(0.8) &91.3(5.8)\\ 
&CIFAR10             &91.4(0.7) &98.4(0.1) &98.0(0.0) &92.6(0.5) &95.6(0.4) &96.9(0.5) &63.1(1.7) &93.3(0.6) &89.5(5.8)\\ 
&CIFAR100            &91.3(0.4) &98.1(0.1) &97.3(0.1) &92.5(0.5) &95.0(0.5) &96.3(0.6) &64.0(1.6) &93.0(0.5) &88.5(6.1)\\ 
&Uniform             &93.7(1.6) &98.7(0.2) &98.6(0.3) &93.0(0.7) &94.8(0.9) &93.9(1.8) &64.6(2.4) &95.1(1.6) &90.1(6.4)\\ 
&Gaussian            &94.4(1.3) &98.8(0.2) &98.7(0.3) &93.6(0.3) &95.6(0.6) &95.4(1.6) &66.5(3.4) &95.1(1.4) &90.3(6.5)\\ 
\midrule
\multirow{8}{*}{\rotatebox[origin=c]{90}{CIFAR-10}}
&Imagenet(c)         &90.0(0.9) &97.7(0.4) &96.6(0.7) &92.3(0.4) &97.4(0.2) &96.6(0.7) &74.3(2.9) &96.4(0.3) &87.3(10.)\\ 
&Imagenet(r)         &87.3(1.3) &96.9(0.6) &95.5(1.0) &91.2(0.3) &96.8(0.4) &95.4(1.4) &71.7(3.1) &95.6(0.5) &86.0(12.)\\ 
&LSUN(c)             &92.0(1.7) &99.0(0.0) &98.8(0.0) &93.7(0.5) &98.7(0.0) &98.7(0.1) &79.9(3.1) &98.4(0.0) &92.0(5.7)\\ 
&LSUN(r)             &91.6(1.2) &98.2(0.4) &96.1(1.2) &93.5(0.3) &98.1(0.1) &95.8(1.3) &72.8(2.9) &97.6(0.2) &85.2(14.)\\ 
&iSUN                &90.1(1.4) &98.0(0.4) &96.2(1.1) &92.9(0.4) &97.9(0.1) &96.0(1.2) &71.0(3.7) &97.3(0.3) &86.0(13.)\\ 
&SVHN                &87.7(2.4) &98.3(0.4) &99.3(0.3) &91.7(0.7) &97.5(0.8) &99.0(0.3) &80.6(5.1) &98.6(0.5) &92.5(4.3)\\ 
&Uniform             &85.9(10.) &93.5(1.0) &97.7(1.6) &88.6(2.5) &99.2(0.6) &93.3(9.0) &75.1(14.) &99.6(0.1) &87.2(5.2)\\ 
&Gaussian            &89.9(10.) &94.6(1.4) &98.3(0.9) &89.6(5.1) &99.2(0.5) &88.6(15.) &77.8(3.4) &99.6(0.2) &85.1(1.5)\\ 
\midrule
\multirow{8}{*}{\rotatebox[origin=c]{90}{CIFAR-100}}
&Imagenet(c)         &78.9(0.1) &83.2(0.6) &63.7(6.6) &76.0(1.5) &93.4(0.7) &57.0(11.) &64.6(0.3) &92.6(0.8) &46.3(10.)\\ 
&Imagenet(r)         &75.1(0.8) &76.6(1.4) &50.1(9.1) &72.0(1.8) &95.5(0.6) &46.6(15.) &61.8(1.5) &91.8(1.1) &51.3(14.)\\ 
&LSUN(c)             &78.8(0.6) &91.3(0.5) &85.7(1.4) &77.5(0.3) &90.1(0.6) &78.1(3.0) &60.5(1.0) &93.3(0.7) &35.6(1.4)\\ 
&LSUN(r)             &76.2(1.4) &78.4(2.5) &46.0(10.) &71.3(0.7) &95.5(0.8) &43.0(14.) &64.1(1.8) &92.0(0.7) &44.6(12.)\\ 
&iSUN                &75.2(1.4) &76.6(2.0) &45.7(10.) &71.4(1.1) &95.2(0.7) &40.0(15.) &61.9(1.9) &91.6(0.8) &45.0(14.)\\ 
&SVHN                &75.1(2.5) &89.6(2.0) &87.9(3.3) &77.5(2.2) &91.5(1.7) &75.3(7.8) &60.0(5.2) &93.6(1.3) &54.3(4.8)\\ 
&Uniform             &69.0(13.) &50.4(11.) &46.8(26.) &84.0(10.) &99.8(0.1) &25.0(17.) &59.2(36.) &99.6(0.1) &97.7(0.8)\\ 
&Gaussian            &51.5(1.8) &31.9(17.) &27.0(19.) &84.8(5.7) &99.9(0.0) &7.75(4.2) &23.5(14.) &99.3(0.3) &99.4(0.4)\\ 
\midrule
\multirow{8}{*}{\rotatebox[origin=c]{90}{CIFAR-100+}}
&Imagenet(c)         &77.0(1.4) &87.0(0.1) &82.1(1.2) &78.2(0.4) &86.8(1.2) &83.1(1.8) &69.4(3.4) &88.3(1.1) &81.6(0.7)\\ 
&Imagenet(r)         &73.7(1.4) &83.8(0.6) &76.5(2.6) &76.3(0.5) &84.3(1.5) &78.0(2.7) &72.4(2.8) &87.0(1.1) &75.4(1.2)\\ 
&LSUN(c)             &77.6(0.5) &89.3(0.5) &89.8(0.8) &76.5(1.1) &90.0(0.3) &90.7(0.7) &55.0(2.3) &86.2(1.3) &89.6(1.1)\\ 
&LSUN(r)             &75.4(2.6) &84.6(1.3) &76.8(1.9) &75.8(2.0) &84.5(0.4) &78.0(2.8) &70.1(3.8) &87.0(2.2) &74.8(2.2)\\ 
&iSUN                &74.5(1.6) &83.6(0.8) &76.2(2.4) &74.9(1.1) &84.0(1.1) &77.7(2.0) &67.9(3.8) &85.7(1.7) &74.5(1.8)\\ 
&SVHN                &72.2(5.9) &83.2(2.9) &81.4(5.2) &74.5(3.4) &86.0(2.6) &83.8(3.4) &67.8(2.4) &86.3(2.7) &84.9(2.2)\\ 
&Uniform             &87.0(1.5) &85.1(6.1) &75.6(14.) &88.7(3.2) &95.0(2.1) &83.6(11.) &60.0(15.) &81.7(14.) &57.3(2.4)\\ 
&Gaussian            &87.1(4.7) &84.2(10.) &81.6(14.) &77.7(4.3) &95.4(4.7) &85.1(10.) &70.0(2.2) &75.9(4.6) &34.9(4.6)\\ 
\bottomrule
\end{tabular}}
%\end{sc}
\end{small}
%\end{center}
\end{table}
%\end{sidewaystable}
\begin{table}[h]
\caption{The summary of classifiers analyzed in the experiment section. Their in-domain classification accuracy is provided in the right four columns. The "+" means that the classifier is trained with extra regularization (dropout rate 0.7).}
\begin{small}
\label{tbl:IDACC}
\begin{center}
\centering
%\resizebox{0.51\textwidth}{!}{
\begin{tabular}{ccccccccc}
\toprule
Classifier       & Image size     & \#class  & Model & Experiment   & Baseline   & DeConf-I   & DeConf-E   & DeConf-C   \\
\midrule
CIFAR-10          & 32x32          & 10       & DenseNet & Table 1,2      & 95.2$\pm$0.1 & 94.9$\pm$0.1 & 95.0$\pm$0.1 & 95.0$\pm$0.1 \\
CIFAR-10          & 32x32          & 10       & ResNet34 & Figure 3       & 95.2$\pm$0.1 & 95.0$\pm$0.1 & 94.9$\pm$0.1 & 95.1$\pm$0.1 \\
SVHN             & 32x32          & 10       & ResNet34 & Figure 3       & 96.9$\pm$0.1 & 96.8$\pm$0.1 & 96.5$\pm$0.1 & 96.7$\pm$0.1 \\
CIFAR-100         & 32x32          & 100      & DenseNet & Table 1,2; Figure 7      & 77.0$\pm$0.2 & 75.8$\pm$0.4 & 76.4$\pm$0.1 & 75.9$\pm$0.1 \\
CIFAR-100         & 32x32          & 100      & WRN      & Figure 7       & 80.8$\pm$0.1 & 78.3$\pm$0.1 & 78.4$\pm$0.1 & 78.4$\pm$0.1 \\
CIFAR-100         & 32x32          & 100      & ResNet50 & Figure 7       & 78.8$\pm$0.3 & 76.4$\pm$0.1 & 76.5$\pm$0.3 & 76.2$\pm$0.2 \\
CIFAR-100         & 32x32          & 100      & ResNet34 & Figure 4,5,7   & 78.5$\pm$0.2 & 76.0$\pm$0.1 & 76.2$\pm$0.1 & 75.8$\pm$0.2 \\
CIFAR-100         & 32x32          & 100      & ResNet18 & Figure 7       & 77.3$\pm$0.1 & 75.2$\pm$0.2 & 75.8$\pm$0.1 & 75.1$\pm$0.1 \\
CIFAR-100         & 32x32          & 100      & ResNet10 & Figure 7       & 75.0$\pm$0.1 & 73.4$\pm$0.1 & 74.2$\pm$0.1 & 73.5$\pm$0.1 \\
CIFAR-100$^+$        & 32x32          & 100      & ResNet34 & Figure 4   & 78.2$\pm$0.1 & 77.4$\pm$0.3 & 77.2$\pm$0.3 & 77.2$\pm$0.1 \\
\begin{tabular}{@{}c@{}}DomainNet \\ (Real-A)\end{tabular} & \begin{tabular}{@{}c@{}}180x180 to \\ 640x880\end{tabular}               & 173   & ResNet34 & Table 3    & 73.6$\pm$0.1 & 73.0$\pm$0.1   & 73.4$\pm$1.5 & 72.2$\pm$0.5 \\
\bottomrule
\end{tabular}
%}
\end{center}
\end{small}
\end{table}

\fi

\end{document}